\documentclass[sigconf]{acmart}
\usepackage{soul}

\usepackage[ruled,lined,commentsnumbered]{algorithm2e}
\usepackage{makecell}
\usepackage{pifont}
\usepackage{multirow}
\usepackage{bm}
\usepackage{placeins}

\AtBeginDocument{%
  }

\copyrightyear{2025}
\acmYear{2025}
\setcopyright{cc}
\setcctype{by}
\acmConference[ISCA '25]{Proceedings of the 52nd Annual International Symposium on Computer Architecture}{June 21--25, 2025}{Tokyo, Japan}
\acmBooktitle{Proceedings of the 52nd Annual International Symposium on Computer Architecture (ISCA '25), June 21--25, 2025, Tokyo, Japan}\acmDOI{10.1145/3695053.3731063}
\acmISBN{979-8-4007-1261-6/2025/06}

\begin{document}

\title{Bishop: Sparsified \underline{B}undl\underline{i}ng \underline{S}piking Transformers on \underline{H}eter\underline{o}geneous Cores with Error-Constrained \underline{P}runing} 

\newcommand{\arch}{\texttt{Bishop}}
\newcommand{\BSA}{\texttt{BSA}}
\newcommand{\ECP}{\texttt{ECP}}

\author{Boxun Xu} 
\affiliation{%
  \institution{Department of Electrical and Computer Engineering, University of California, Santa Barbara}
  \city{Santa Barbara}
  \state{CA}
  \country{USA}
}
\author{Yuxuan Yin}
\affiliation{%
  \institution{Department of Electrical and Computer Engineering, University of California, Santa Barbara}
  \city{Santa Barbara}
  \state{CA}
  \country{USA}
}
\author{Vikram Iyer}
\affiliation{%
  \institution{Department of Electrical and Computer Engineering, University of California, Santa Barbara}
  \city{Santa Barbara}
  \state{CA}
  \country{USA}
}
\author{Peng Li}
\affiliation{%
  \institution{Department of Electrical and Computer Engineering, University of California, Santa Barbara}
  \city{Santa Barbara}
  \state{CA}
  \country{USA}
}

\begin{abstract}
Spiking neural networks(SNNs) have emerged as a promising solution for deployment on resource-constrained edge devices and neuromorphic hardware due to their low power consumption. Spiking transformers, which integrate attention mechanisms similar to those found in artificial neural networks (ANNs), have recently exhibited impressive performance. However, these models are large in size and involve high-volume computation in both time and space, posing significant challenges for efficient hardware acceleration. We present \arch,
the first dedicated hardware accelerator architecture and HW/SW co-design framework for spiking transformers that optimally represents, manages, and processes spike-based workloads while exploring spatiotemporal sparsity and data reuse. Specifically, we introduce the concept of Token-Time Bundle (TTB), a container that bundles spiking data of a set of tokens over multiple time points. Our heterogeneous accelerator architecture \arch \space concurrently processes workload packed in TTBs and explores intra- and inter-bundle multiple-bit weight reuse to significantly reduce memory access. 
\arch \space utilizes a stratifier, a dense core array, and a sparse core array to process MLP blocks and projection layers. The stratifier routes high-density spiking activation workload to the dense core and low-density counterpart to the sparse core, ensuring optimized processing tailored to the given spatiotemporal sparsity level. To further reduce data access and computation, we introduce a novel Bundle Sparsity-Aware (BSA) training pipeline that enhances not only the overall but also structured TTB-level firing sparsity. Moreover, the processing efficiency of self-attention layers is boosted by the proposed Error-Constrained TTB Pruning (\ECP), which trims activities in spiking queries, keys, and values both before and after the computation of spiking attention maps with a well-defined error bound. Finally, we design a reconfigurable TTB spiking attention core to efficiently compute spiking attention maps by executing highly simplified ``AND” and ``Accumulate” operations. On average, \arch \space achieves a $5.91\times$ speedup and $6.11\times$ improvement in energy efficiency over previous SNN accelerators, while delivering higher accuracy across multiple datasets.
\end{abstract}



\maketitle

\section{Introduction}
As the third generation of neural networks \cite{maass_snn_1997}, spiking neural networks (SNNs) exhibit a closer resemblance to biological neural circuits than their non-spiking artificial neural network (ANN) counterparts. SNNs hold promise owing to their biological plausibility, event-driven characteristics, and low power consumption~\cite{RoyNature:2019}.
The recent spiking-based transformer models have demonstrated superior performance compared to conventional spiking network architectures \cite{spikformer, spikformer_tracking, bal2024spikingbert, zhu2023spikegpt, yao2023spike}, mirroring the trend observed in ANNs where vision transformers outperform VGGs or ResNets. 
A number of hardware accelerators for ANN-based transformers have been proposed.  For example, \cite{qu_dota_2022, dao2022flashattention} focus on more efficient processing of attention-based computations; \cite{you_castling-vit_2023,fan_adaptable_2022} propose techniques dealing with transformers exhibiting a targeted level of sparsity; \cite{dass_vitality_2023} utilizes Taylor approximation to eliminate the complex non-local softmax functions in attention computation.

However, hardware acceleration of spiking transformers remains largely unexplored, giving rise to significant challenges related to computational overhead, latency, and energy consumption, especially when dealing with large spiking transformers. 
Continuing the use of hardware architectures designed for spiking fully connected (FC) or convolutional neural networks (CNNs) or adapting existing ANN-based transformer accelerators \cite{wang2021spatten, lu2021sanger, zhao2024alisa} falls short in harnessing the unique characteristics of spiking transformers. 

\begin{figure*}[ht]
    \centering
        \includegraphics[width=\textwidth, clip, trim={3.5cm 9.5cm 3cm 2cm}]{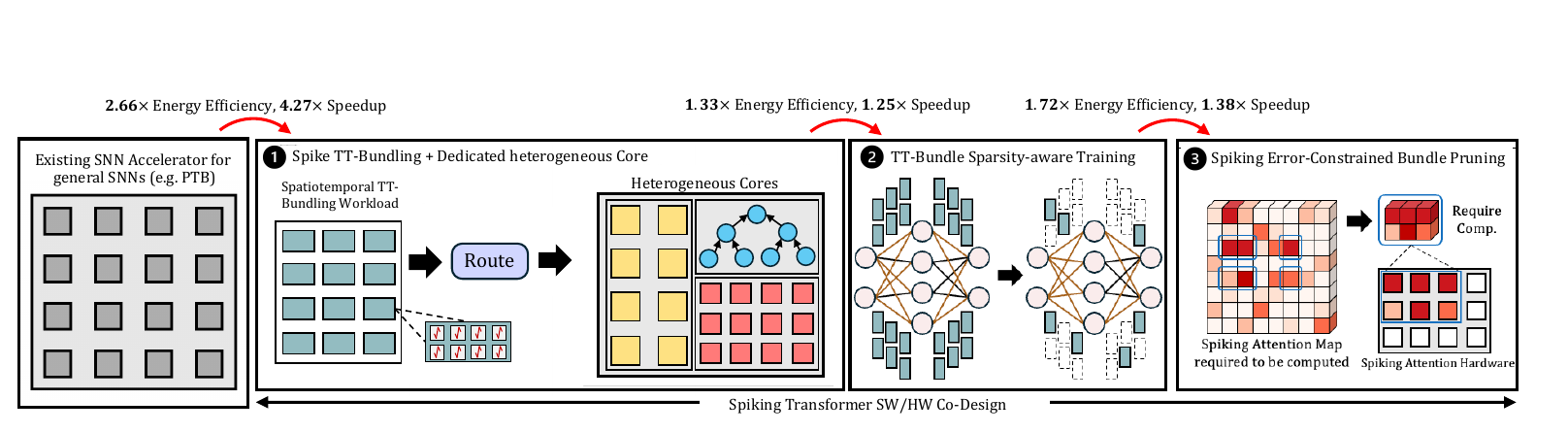}
    \caption{{\arch}: the first accelerator architecture and SW/HW co-design framework dedicated to spiking transformers.}
    \Description{}
    \label{fig:proposed}
\end{figure*}

This work presents the first dedicated hardware accelerator architecture and HW/SW co-design framework tailored for spiking transformers. We contend that the most promising approach to accelerate spiking transformer models is to optimally represent, manage, and process spike-based workloads while exploring spatiotemporal sparsity and data reuse. As illustrated in Fig.~\ref{fig:proposed}, this work makes the following contributions.

In the context of spiking vision transformers, we introduce the concept of spiking Token-Time Bundles (TTBs). A TTB is a container that bundles spiking data for a set of spatial tokens over a number of time points, capturing the temporal variations of these tokens. The TTB serves as the fundamental unit of work to be processed on our proposed \arch \space architecture. The utilization of TTBs empowers us to harness structured data reuse and firing sparsity inherent in key computations within the transformer across both time and space. 

Our \arch\space architecture \ding{182} is heterogeneously comprised of a TTB stratifier, a TTB dense core, a TTB sparse core, and a TTB attention core. The TTB stratifier efficiently routes high-density spiking-TTB workloads to the TTB dense core and directs low-density workloads to the TTB sparse core, ensuring optimized processing tailored to the present spatiotemporal sparsity level of the workload. Taking advantage of the binary nature of spiking TT-bundled queries and keys, we design a reconfigurable multiplier-less And-ACcumulate (AAC) array. This array efficiently processes spiking attention layers by executing highly simplified ``AND" and ``Accumulate" operations and optimized dataflow, achieving a $2.66\times$ improvement in energy efficiency and a $4.27\times$ speedup over the Parallel Time Batching (PTB) architecture of \cite{PTB}.

From the algorithmic perspective, we explore firing sparsity tailored for hardware as a key opportunity for efficient acceleration. To this end, we introduce \ding{183}, a TTB sparsity-aware training pipeline that not only enhances the overall sparsity of spiking activations but does so in a structured manner by organizing them into TTBs for efficient processing on \arch. Furthermore, we propose \ding{184}, an Error-Constrained TT-Bundle Pruning (\ECP) algorithm to eliminate redundant operations across tokens of spiking queries, keys, scores, and values with a well-defined pruning threshold, facilitating efficient attention computation in conjunction with \ding{182} while maintaining high attainable performance.

\section{Background}

\subsection{Spiking Neural Networks}
\textbf{Leaky Integrate-and-Fire (LIF) Model.} We adopt the widely used leaky integrate-and-fire neuronal model in SNNs \cite{gerstner2002spiking}, which has the following discretized dynamics over time:
\begin{eqnarray}
V_{m}\left[t_{k}\right] = V_{m}\left[t_{k-1}\right] + I[t_{k}] - V_{leak} \label{eqn_LIF_1} \\
S\left[t_{k}\right] = \begin{cases} 
1 & \text{if  } V_{m}[t_{k}] > V_{th} \rightarrow V_{m}[t_{k}]=0\\
0 & \text{else          }       \rightarrow V_{m}[t_{k}] = V_{m}[t_{k}] 
\end{cases} \label{eqn_LIF_2}
\end{eqnarray}
where $V_{m}\left[t_{k}\right]$ is the membrane potential of a spiking neuron at time point $t_{k}$; $I[t_{k}]$ is the total synaptic input current at $t_{k}$; $V_{leak}$ is the leakage; $V_{th}$ is the firing threshold, and $S\left[t_{k}\right]$ is the spiking output. A spiking output is generated if the membrane potential $V_{m}\left[t_{k}\right]$ exceeds $V_{th}$, setting $S\left[t_{k}\right]$ to 1 and resetting $V_{m}[t_{k}]$ to 0.  

\begin{figure}[hb]
    \centering
    \includegraphics[width=0.99\columnwidth]{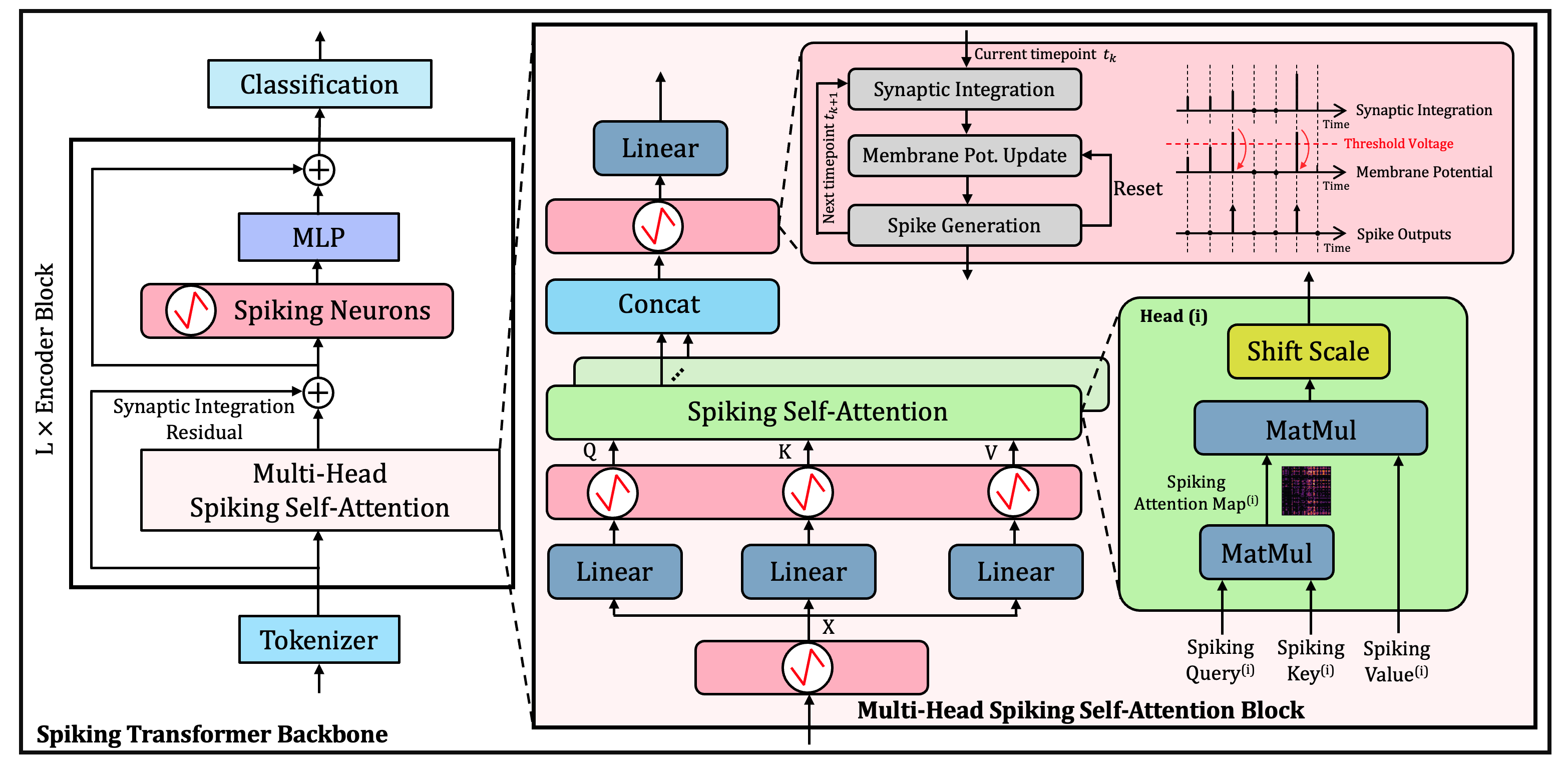}
    \caption{Spiking transformer model architecture with multi-head spiking self-attention \cite{yao2024spike, spikformer}.}
    \Description{}
    \label{fig:MultiHead}
\end{figure}

\textbf{Spiking Transformers.}
The recent emergence of spiking transformers presents a major development in spiking neural networks \cite{spikformer, spikformer_tracking, zhu2023spikegpt, yao2023spike, wang2023masked},  showing large performance improvements over other SNNs \cite{roy2019towards, sengupta2019going} such as spiking CNNs \cite{lee2020enabling, rathi2020enabling, li2021differentiable, fang2021deep}. Specifically, the state-of-the-art spiking transformers incorporate a multi-head Spiking Self-Attention (SSA) mechanism \cite{yao2023spike, yao2024spike}, capturing global correlations among input tokens in a spike form. This approach demonstrates impressive performance with promising energy efficiency in vision applications compared with ANNs under the same model size. Tab.~\ref{table:ann_snn_acc_benchmark} compares the accuracy of spiking transformers with other state-of-the-art ANNs and SNNs across various image\cite{deng2009imagenet}, gesture\cite{amir2017low} and speech command\cite{warden2018speech} recognition datasets. 

As depicted in Fig.~\ref{fig:MultiHead}, spiking transformers incorporate a spiking tokenizer, which transforms a static image or a dynamic input, e.g, a dynamic vision sensor (DVS) stream, represented as an input sequence  $I \in\mathbb{R}^{T\times C\times H\times W}$ with spatial size $H\times W$ in $C$ channels across $T$ time points, into  $I' \in\mathbb{R}^{T\times N\times D}$, i.e., $N$ $D$-dimensional tokens across $T$ time points. The core model architecture comprises $L$ sequentially connected residual encoder blocks with each consisting of a multiple-head Spiking Self-Attention (SSA) block and a spiking MLP block. The output from the final encoder block is processed with a global average pooling across all tokens and time points, and is subsequently employed to make the final prediction through a linear layer\cite{dosovitskiy2020image}.

\textbf{Spiking Self-Attention.}\label{sec:spikingattention}
Unlike conventional spiking CNNs, spiking transformers feature the use of SSA blocks for extracting  global token dependencies. 
The SSA blocks in our proposed {\arch} architecture operate as follows: 
\begin{eqnarray}
    Q^{(i)} &=& \mathcal{LIF}(X^{(i)} \cdot W^{(i)}_{Q}) \\
    K^{(i)} &=& \mathcal{LIF}(X^{(i)} \cdot W^{(i)}_{K}) \\
    V^{(i)} &=& \mathcal{LIF}(X^{(i)} \cdot W^{(i)}_{V}) \\
    O^{(i)} &=& ({Q^{(i)} \cdot K^{(i)T}} \cdot s) \cdot V^{(i)} \label{eq:attention} \\ 
    O_{temp} &=& \mathcal{LIF}(Concat\{O^{(1)}, ... ,O^{(H)}\}) \label{eq:lif_concat} \\
    O_{attn} &=& (O_{temp} \cdot W_{O}), \label{eq:lifattout}
\end{eqnarray}
where $X$ represents the binary inputs produced by the preceding LIF neurons, and $H$ denotes the number of heads. For each $i$-th head, $Q^{(i)}$,  $K^{(i)}$,  $V^{(i)}$, $O^{(i)}$ represent the query, key, value, and output respectively; $W^{(i)}_{Q}$, $W^{(i)}_{K}$, $W^{(i)}_{V}$ are the weights of the corresponding linear projection layers for computing $Q^{(i)}$, $K^{(i)}$, and $V^{(i)}$, respectively. $\mathcal{LIF}$ refers to an LIF neuron layer. $s$ is a power-of-two scaling parameter, allowing for efficient scaling via bit shifting. $W_{O}$ represents the weights of the final linear projection layer, and $O_{attn}$ is the output from the $H$-head SSA block. 
In contrast to the method described in \cite{spikformer} which extensively employs multipliers due to spike residuals, we reposition the final LIF neuron layer to precede the last linear layer of each SSA block\cite{yao2024spike} as in (\ref{eq:lif_concat}). This adjustment allows for efficient multiplication-free computation of attention output $O_{attn}$ based on the product of the weights $W_o$ with the binary LIF activations $O_{temp}$ in (\ref{eq:lifattout}).

\begin{table}[tbp]
\centering
\small
\caption{Comparison of ANN and SNN Accuracies.}
\setlength{\tabcolsep}{0.5pt} 
\begin{tabular}{ccc}
\toprule
\textbf{Workload} & \textbf{ANN Accuracy} & \textbf{SNN Accuracy} \\
\midrule
\multirow{5}{*}{CIFAR10} 
    & ResNet-19\cite{deng2022temporal}: 94.97\%    & VGG-11\cite{rathi2020enabling}: 92.22\% \\
    & Transformer\cite{dosovitskiy2020image}: 96.73\% & CIFARNet\cite{TSSLBP}: 91.41\% \\
    &  & ResNet-19\cite{zheng2021going}: 92.92\% \\
    &                      & ResNet-20\cite{rathi2021diet}: 92.54\% \\
    &                      & Spiking Transformer: \textbf{95.19\%} \\
\midrule
\multirow{4}{*}{CIFAR100} 
    & ResNet-19\cite{deng2022temporal}: 75.35\%  & VGG-11\cite{rathi2020enabling} 67.87\% \\
    & Transformer\cite{dosovitskiy2020image}: 81.02\%& ResNet-19\cite{zheng2021going}: 70.86\% \\
    &   & ResNet-20\cite{rathi2021diet}: 64.07\% \\
    &                     & Spiking Transformer: \textbf{77.86\%} \\
\midrule
\multirow{3}{*}{DVS-Gesture} 
    & 12-layer CNN: 94.59\%\cite{amir2017low} & CIFARNet \cite{TSSLBP}: 91.32\% \\
    &   & ResNet-19\cite{zheng2021going}): 96.9\% \\
    &   & Spiking Transformer: \textbf{98.3\%} \\
\midrule
\multirow{3}{*}{ImageNet} 
    & ResNet-34\cite{sengupta2019going}: 70.69\% & Spiking ResNet-34\cite{deng2022temporal}: 64.79\% \\
    & Transformer\cite{dosovitskiy2020image}: 80.8\% & SEW-ResNet-50\cite{fang2021deep}: 67.78\% \\
    &                     & Spiking Transformer: \textbf{73.38\%} \\
\midrule
    \multirow{2}{*}{\color{black} Google SC}
    & {\color{black} ConvNet\cite{jansson2018single}: 87.0\%} & {\color{black} Spiking ResNet\cite{yang2022deep}: 92.9\%} \\
    & {\color{black} AttentionRNN\cite{de2018neural}: 93.9\%} & {\color{black} Spiking Transformer: \textbf{95.11\%}} \\
\bottomrule
\end{tabular}
\label{table:ann_snn_acc_benchmark}
\end{table}

\subsection{Complexity/Workload Profiling of Spiking Transformers}\label{subsection: profiling}
\textbf{Computational Complexity.}\label{sec:complexity}
We analyze the computational complexity of a spiking transformer. As shown in Fig.~\ref{fig:MultiHead}, the MLP block, the four ($Q$, $K$, $V$, $O$) linear projection layers, and attention layers are mostly dominant sources of complexity. In the MLP block and linear projection layers of a multi-head SSA block, the spike inputs are multiplied with the corresponding weights to produce the synaptic inputs to the LIF neuron units, which generate firing outputs. The space and time complexity of these layers is $O(TND^2)$. The attention layers within an attention block compute the dot-product of spiking queries and keys with a computational complexity of $O(TN^2D)$. With $N \gg D$, the complexity of the attention layers dominates that of the MLP/projection layers. The opposite is true with $D \gg N$. The LIF neuron layers have a non-dominant   complexity of $O(TND)$. The tokenizer's computational complexity is $O(THWC^2K^2)$, where $K$ is the size of the employed spiking convolutional (\texttt{CONV}) filters. It is typically not the dominant complexity, and there has been much work targeting hardware acceleration of spiking CNNs \cite{Spinalflow, PTB}.

\begin{figure}[H]
    \centering
    \includegraphics[width=\columnwidth, clip, trim={3.6cm 7cm 1.3cm 2.5cm}]{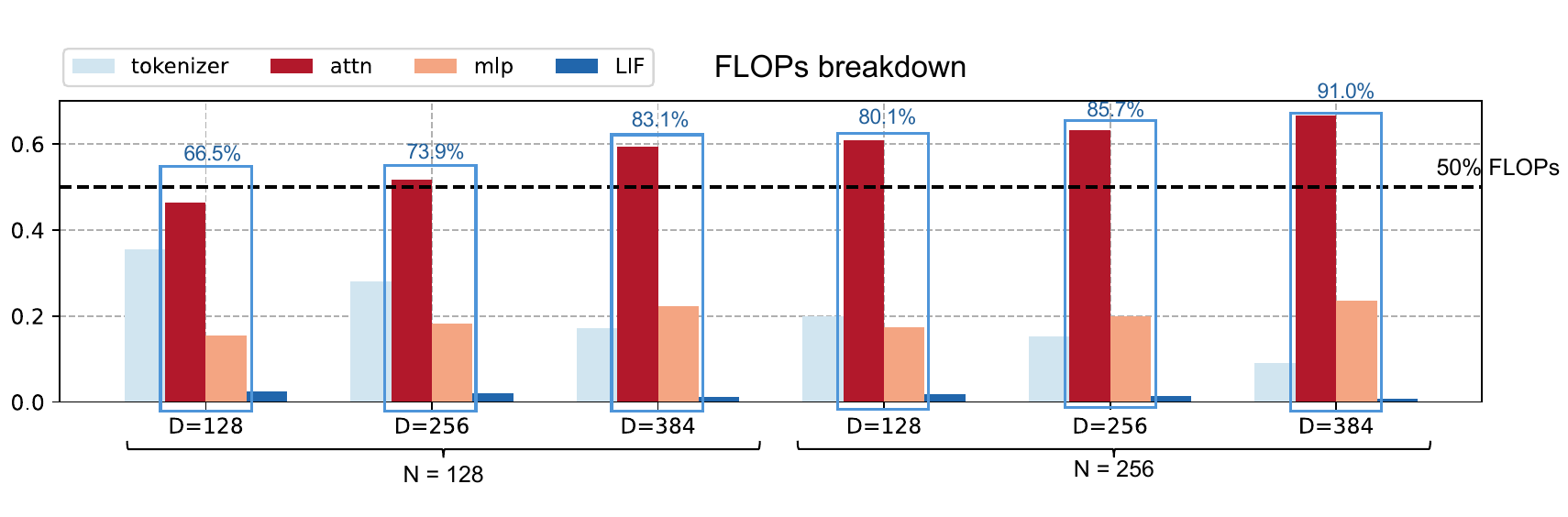}
    \caption{The FLOPs breakdown of a spiking transformer with different token and feature sizes trained on ImageNet.}
    \Description{}
    \label{fig:profiling}
\end{figure}

\textbf{Workload Profiling.}
To shed more light on the inference complexity of spiking transformers, we perform profiling and provide a FLOP breakdown for a spiking transformer based on the architecture of \cite{spikformer} and trained on the ImageNet dataset. Fig.~\ref{fig:profiling} reveals that the primary computational burdens reside within the spiking attention (Attn) and MLP blocks,  especially in deeper spiking transformers. In addition, the dominance of attention blocks over MLP blocks intensifies as $N$ increases. The cumulative FLOPs attributed to these blocks range from 66.5\% to 91.0\% of the total workload. Hence, attention and MLP blocks are the primary target of the proposed accelerator architecture. 

\subsection{Existing SNN accelerators}\label{sec:existing_snn}

\begin{figure*}[ht]
    \centering
    \includegraphics[width=0.99\textwidth, clip, trim={6.5cm, 4.5cm, 1cm, 1cm}]{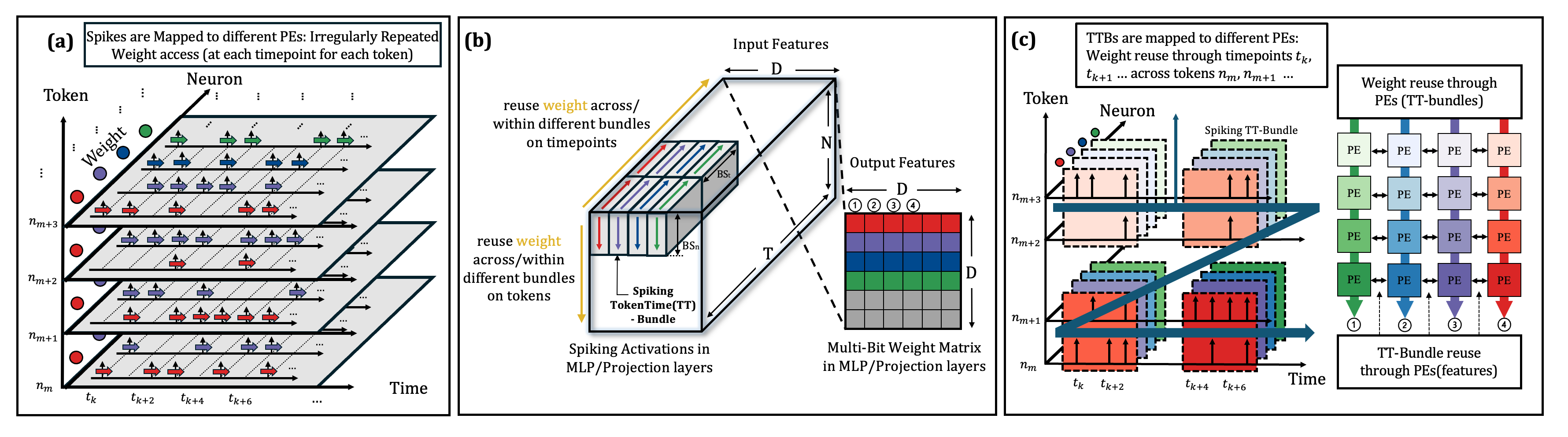}
    \caption{(a) The conventional approach lacks parallel processing in time and space, processing each token at each timepoint in a time-serial manner, causing irregularly repeated weight accesses. (b) The spiking Token-Time (TT)-Bundle allows for multi-bit weight data reuse across multiple tokens and time points. (c) Weight reuse in both an intra- and inter-bundle manner. Multiple TT bundles of neurons corresponding to different output features are mapped onto different PE columns. Multiple TT bundles across different input features marked as different color depths are mapped onto different PE rows.}
    \Description{}
    \label{fig:Bundle_concept}
\end{figure*}

There exists a body of SNN accelerators focusing on enhancing inference energy efficiency or latency through the employment of novel devices \cite{nebula}, circuits \cite{liu202430}, processing architectures \cite{Spinalflow, PTB, Lee_ICCD_2020, mao2024stellar,yin2024loas} and on-chip communication networks \cite{debole2019truenorth, DynamicBundle, davies2018loihi, lee2018flexon, liu2024activen}, or on improving training efficiency  \cite{skipper, yin2023sata, liang2021h2learn}. These efforts predominantly target spiking CNNs such as spiking-based AlexNet/VGG/ResNets \cite{fang2021deep, li2022converting, zheng2021going}. \cite{10454330} extends traditional ANN transformer models by enabling them to operate under both integer and rate-encoded spiking modes. However, these architectures are not optimized for large spiking transformers.

We identify four major challenges and opportunities with respect to hardware acceleration of spiking transformers. 
Firstly, given the excessive use of multi-bit weight data in spiking transformers and the fact that spiking activations are single-bit, it is crucial to maximize weight reuse both temporally and spatially. While spiking CNN accelerators \cite{PTB, Lee_ICCD_2020} attempt to reduce weight data access, they are limited to exploiting weight reuse only in the temporal dimension within a systolic array. Instead, a proper packing of the spatiotemporal spiking workloads optimized for the computational structure within spiking transformers can lead to structured sparsity and improve weight reuse across both time and space. 
Secondly,  existing SNN accelerators \cite{Spinalflow, PTB, nebula, skipper} do not explore a heterogeneous architecture, leading to large efficiency loss when encountering workloads with a varying degree of structured or unstructured sparsity. 
Thirdly, the high performance of recent spiking transformers can be largely attributed to the new spiking attention mechanisms adopted \cite{spikformer, spikformer_tracking, bal2024spikingbert, zhu2023spikegpt, yao2023spike}. Nonetheless, dedicated architectural support for these mechanisms remains largely absent.
Finally, while algorithm/hardware co-optimization such as weight pruning \cite{STDP-Based-Pruning, chen2021pruning}, temporal pruning \cite{chowdhury2022towards}, and quantization \cite{yin2024mint} have demonstrated effectiveness for spiking CNNs, a systematic approach to accelerator architecture design and algorithm/hardware co-optimization is currently lacking for spiking transformers. 
Focusing on 3D physical design, \cite{xu2024spiking} proposes a simple spiking transformer architecture that does not pack spatiotemporal tokens into bundles, thereby missing opportunities for effective weight data reuse and for supporting temporal and token-level parallel processing within each processing unit, nor does it exploit any form of spiking sparsity or adequately consider HW/SW co-design.

To this end, we propose \arch, which to the best of our knowledge is the first accelerator architecture and HW/SW co-design framework dedicated to spiking transformers.  It's designed to fully exploit irregular sparse firing patterns and maximize data reuse on a heterogeneous array architecture. Additionally, it incorporates a sparsity-aware training pipeline and error-constrained pruning tailored for spiking transformers. These approaches significantly reduce processing latency and enhance energy efficiency, effectively addressing critical computational and data access bottlenecks in state-of-the-art spiking transformers. 

\section{Spiking Token-Time (TT) Bundle}

\subsection{Challenges in Managing Spatiotemporal Workloads}
Spatiotemporal workloads should be appropriately packed to facilitate efficient processing.  
In \cite{Lee_ICCD_2020} a spiking systolic array is proposed to spatially process different postsynaptic neurons in parallel and temporally process the input integration of each neuron over several sequential time steps. However, the lack of time-parallel processing exacerbates the overhead of accessing expensive multiple-bit, e.g., 8 bits,  weight data. The systolic-array-based Parallel Time Batching (PTB) architecture in \cite{PTB} addresses this issue by packing spiking activities across multiple time points within a time window and processing several such windows concurrently on the systolic array, thereby improving weight data reuse.

However, both works only support the computations of convolution neural networks (CNNs) and fully-connected layers (FCs) without addressing the unique computation structures of spiking transformers. 
For instance, FC layers perform matrix-vector multiplications on spiking activtations with a vector length  equal to the number of neurons. On the other hand, the main computations in spiking transformers have a very different spatiotemporal computation structure. MLP layers in transformers perform matrix-matrix multiplications on spiking activations in the dimension of N (tokens) $\times$ D (features) across T time points based on weights whose dimension is $D\times D$. Furthermore, SNNs may operate over a wide range of timesteps. Packing workloads solely in time as in the PTB architecture \cite{PTB} is only effective when the number of timesteps is large, e.g., 100-300. The number of operational timesteps of spiking transformers may vary from 
300 \cite{wang2023masked} down to 4 \cite{spikformer,spikformer_tracking}. 
Operating with a reduced number of time steps significantly reduces both inference latency and training costs, marking a trend driven by advances in SNN training algorithms. To maintain flexibility in hardware architectures, there is a need for supporting spatiotemporal computation within spiking transformers across a wide range of time horizons, with a specific emphasis on facilitating computation over shorter time spans.

\subsection{Proposed Spiking Token-Time (TT) Bundle}
The most promising approach to accelerate spiking transformer models is to optimally represent, manage, and process spike-based workloads while exploring spatiotemporal sparsity and opportunities for data reuse. We introduce the concept of spiking Token-Time Bundle (TTB), a container that bundles spiking data for a set of tokens over a set of time points, specifically for spiking transformers. As shown in Fig.~\ref{fig:Bundle_concept}, we split the spiking activities of $N$ tokens, $D$ features, across $T$ time points into multiple token-time bundles. Each TTB packs $BS_{n}$  tokens across $BS_{t}$ timepoints for a given output feature. Accordingly, we have $\lceil T/BS_{t} \rceil \times \lceil N/BS_{n} \rceil$ token-time bundles. The utilization of TTBs provides two major  benefits in data reuse and processing.  

\subsubsection{Multi-bit Weight Data Reuse}
Without data packing in TTBs, an accelerator irregularly and repeatedly loads weight data to process individual input spikes of a token at a  time point, discarding much of the opportunities in data reuse, as shown in Fig.~\ref{fig:Bundle_concept}(a),  where processing of each spike is mapped onto a PE. In contrast, the proposed {\arch} architecture maps the processing of a TTB onto a PE, and exploits weight data reuse at multiple levels as shown in Fig.~\ref{fig:Bundle_concept}(b). Because the same multi-bit weights are used for different time points and different tokens for each feature, the $1\times D$ weight data in one row, for example, the highlighted red row, is shared within each TTB for processing $BS_{n}$  tokens across $BS_{t}$ time points. In addition, the same weight data is reused across $\lceil T/BS_{t} \rceil \times \lceil N/BS_{n}\rceil$ TTBs. As such, {\arch}  simultaneously exploits both intra-TTB and inter-TTB weight data reuse. In addition, the same input activiations are shared within each PE row. 

\subsubsection{Explore structured TTB-Level Sparsity}
Furthermore, we tag all TTBs and categorize them into two groups. A TTB is called \emph{active} if there exists at least one active spike in the bundle. Otherwise, we call it \emph{inactive}. Inherent spatiotemporal sparsity of spiking activities can be leveraged to skip processing inactive TTBs, which constitute of a bulk of all bundles. Since {\arch} packs and dispatches workloads in terms of TTBs, skipping inactive bundles can be efficiently accommodated in the dataflow and avoids large overheads resulting from skipping computations at a much smaller granularity, e.g., at the spike or token level. The effectiveness of this approach is further enhanced by the Bundle-Sparsity Aware Training (BSA) algorithm discussed in Section~\ref{sec:ba_training}. 

\section{Proposed Spiking Transformer Algorithms}
We present two algorithms, namely spiking TT Bundle-Sparsity Aware Training (\BSA) and Error-Constrained TT Bundle Pruning (\ECP) to optimize bundle-level sparsity and prune spiking activities in queries and keys, respectively. \ECP\space is also integrated into the training pipeline, leading to \ECP-aware training to maintain high accuracy. \BSA\space and  \ECP\space provide an end-to-end algorithm and hardware co-optimization approach for the proposed {\arch} architecture. 

\subsection{Bundle-Sparsity Aware Training (\BSA)}\label{sec:ba_training}

\begin{figure}[htbp]
    \centering
    \includegraphics[width=0.99\columnwidth, clip, trim={9cm, 11cm, 9cm, 1cm}]{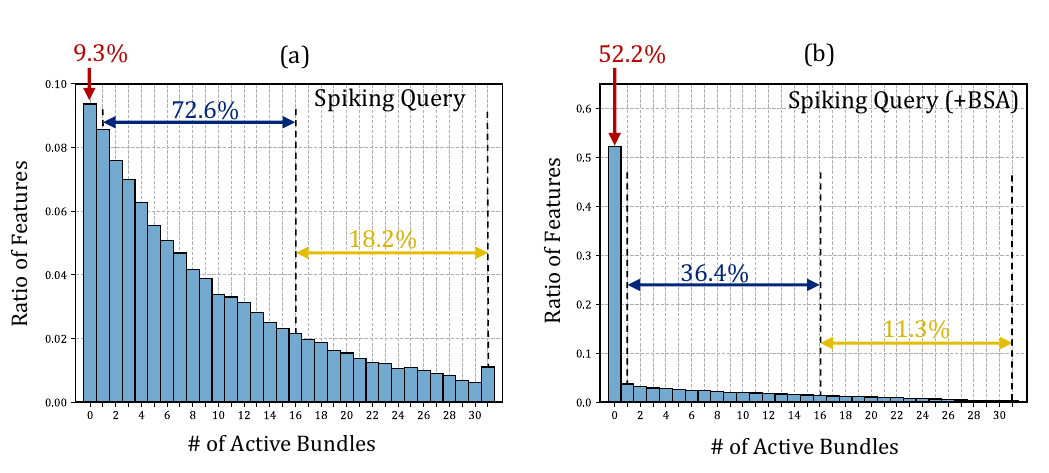}
    \caption{Active bundle distribution of spiking queries (Q) across each input feature in the $4^{th}$ encoder block of a spiking ViT (Model 1)  trained on CIFAR10: (a) without \BSA, and (b) with \BSA.}
    \Description{}
    \label{fig:BSA_sample}
\end{figure}

We visualize the distribution of active token-time bundles (TTBs) in spiking transformers in Fig.~\ref{fig:BSA_sample}(a). Without applying \BSA, spiking transformers exhibit moderate TTB-level sparsity. For instance, in Model 1 of Table~\ref{table:trans_arch}, 29\% of the bundles are active across all layers, offering  some restricted opportunities for computation skipping.

Unlike suppressing activities at the individual spike level as in \cite{Activity_Regularization}, the proposed Bundle-Sparsity Aware Training (\BSA) algorithm sparsifies spiking activities at the TTB level, i.e., by reducing the number of active TTBs, thus providing a structured way of exploiting sparsity.

\begin{figure}[htbp]
    \centering
    \includegraphics[width=0.99\columnwidth, clip, trim={4.5cm, 11cm, 6.7cm, 3.5cm}]{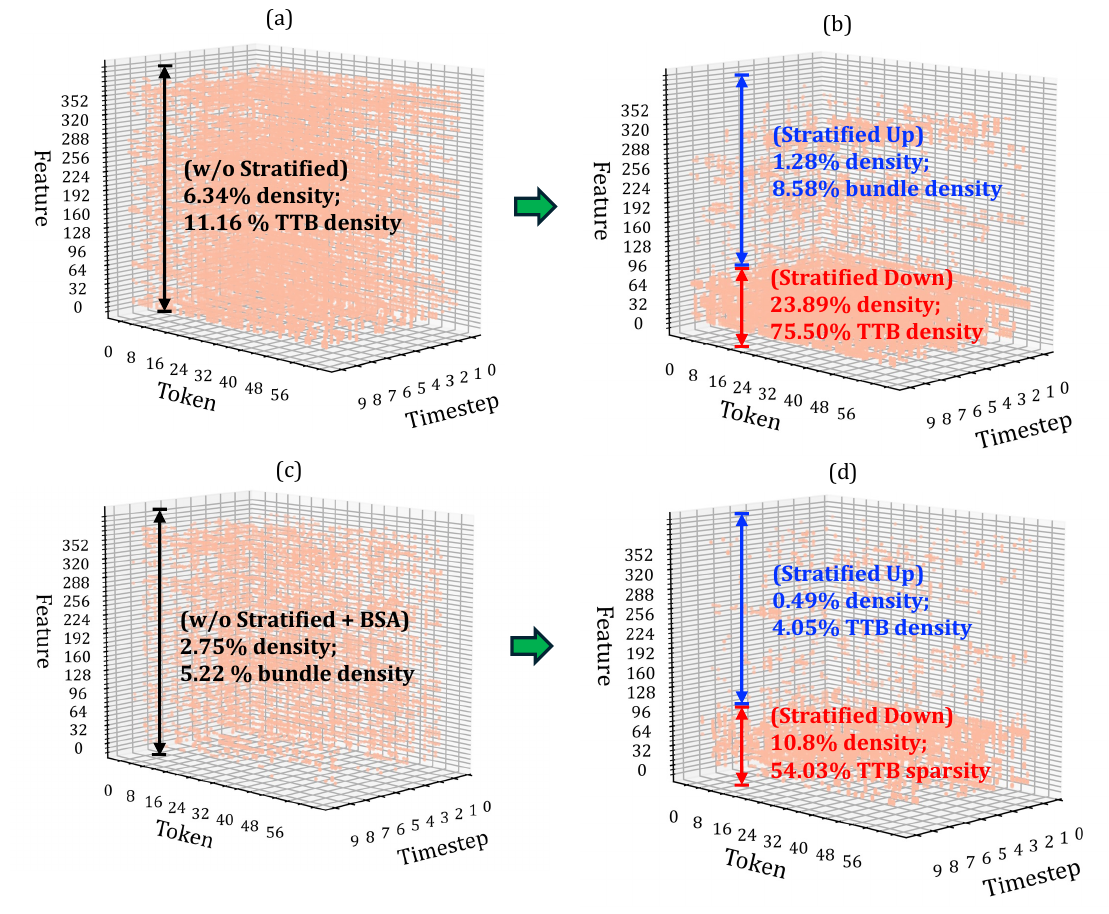}
    \caption{Spiking activities at the output projection layer in the $3^{rd}$ encoder block of a spiking ViT (Model 1) trained on CIFAR10: (a) the original workload without \BSA\space  training, (b) stratified TTBs without \BSA\space, (c) the sparsified workload with BSA, and (d) stratified TTBs with \BSA.}
    \Description{}
    \label{fig:Stratify_vis}
\end{figure}

In \BSA, we denote the spiking activations  at time step $t$ in layer $l$, token $n$, and input feature dimension $d$ by $X^{(l)}_{n,t,d}$, and a bundle for layer $l$  by $TTB^{(l)}_{bn, bt, d}$, where $bn$, $bt$, and $d$ are the bundle-token, bundle-time, and input feature index, respectively. While $TTB^{(l)}_{bn, bt, d}$ packs multiple tokens and time steps, its activity tag  $Z_{bn,bt,d}$  is defined by the  $L_{0}$ norm of the spiking activations  that fall within it: 

\begin{figure*}[ht]
    \centering
    \includegraphics[width=1.0\textwidth, clip, trim={1.5cm 3.2cm 0.9cm 4.3cm}]{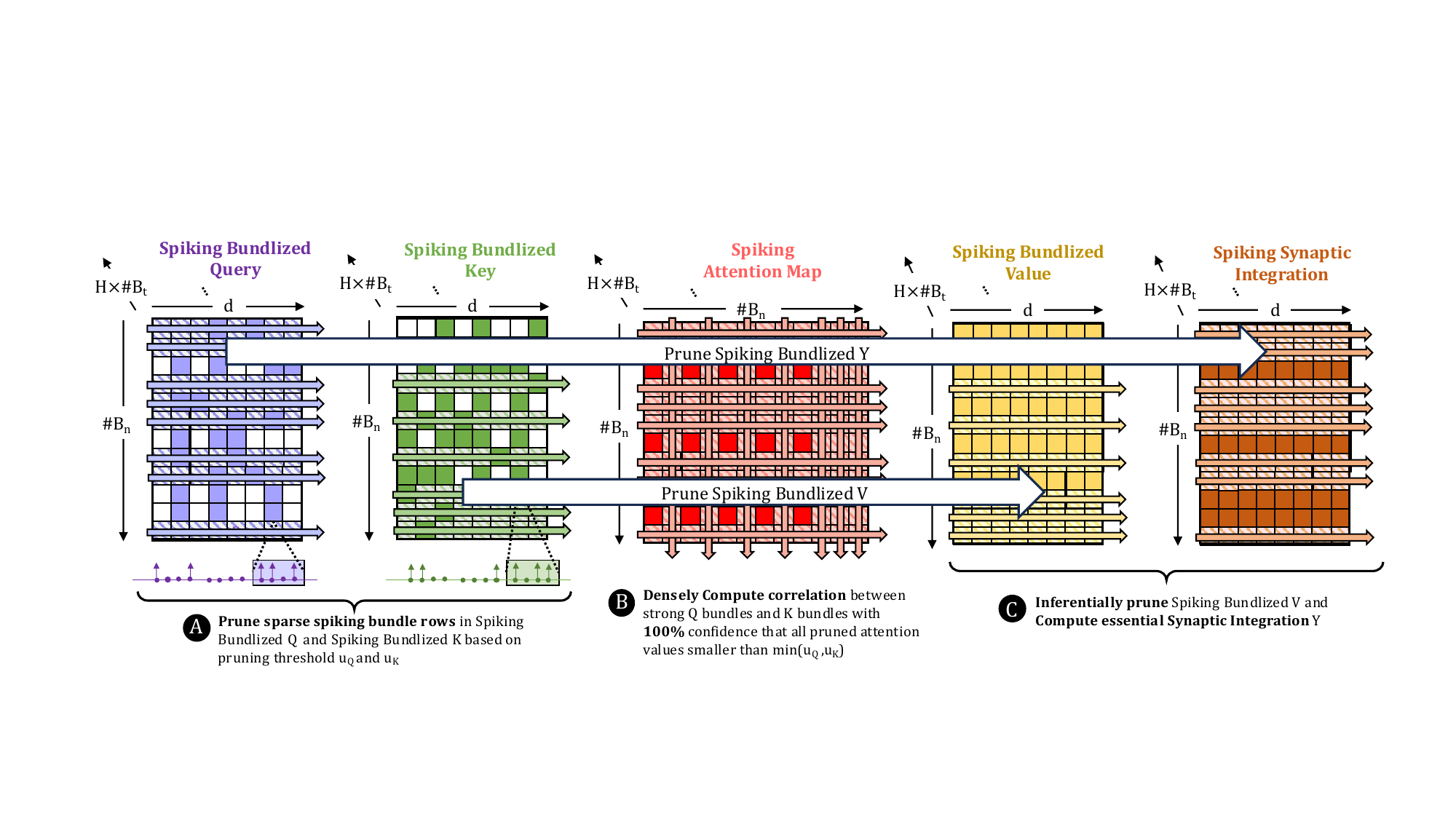}
    \caption{Error-constrained TT bundle pruning (\ECP) for queries (Q), keys (K), attention scores (S), values (V), and outputs (Y).}
    \Description{}
    \label{fig:ECP_illustration}
\end{figure*}

\begin{equation}
\centering
    Z_{bn,bt,d} = ||X_{bn\cdot BS_{n}:(bn+1)\cdot BS_{n}-1, bt \cdot BS_{t}:(bt+1) \cdot BS_{t}-1,d}||_{0}
\label{eq:bundle-tag}
\end{equation}
$Z_{bn,bt,d}$ is zero if there is no   active spike in the bundle. 
We sum up all bundle tags across all $D$ features and $L$ layers of the transformer to define a bundle-level sparsity loss $L_{bsp}$:  
\begin{equation}\label{qn:bsparsity}
\centering
L_{bsp} = \sum_{l =0}^{L-1} \sum_{bn=0}^{\lceil N/BS_{n} \rceil -1} \sum_{bt=0}^{\lceil T/BS_{t} \rceil -1} \sum_{d=0}^{D -1}  Z_{bn,bt,d}
\end{equation}
When forming the above $L_{bsp}$ loss, we consider activations from all MLP and linear projection layers as well as the bundles associated with the queries (Q) and keys(K) in the attention layers.
\BSA\space optimizes the model weight parameters $\bm \theta$ by minimizing a total loss: $L_{tot} = L_{CE} + \lambda L_{bsp}$, which jointly considers a cross-entropy model accuracy loss $L_{CE}$ and $L_{bsp}$ with a hyperparameter $\lambda$, controlling the tradeoff between the accuracy and TTB-level sparsity.

Fig.~\ref{fig:BSA_sample} visualizes the original and \BSA-altered distributions of the active bundles of Model 1, defined in Table~\ref{table:trans_arch}. 
Specifically, \BSA\space significantly improves TTB sparsity and reshapes their distribution across input features, leading to most features having only a small number of active bundles.
In addition, \BSA\space leads to a substantial increase in the percentage of input features (from 9.3\% to 52.2\% in Model 1, as shown in Fig.~\ref{fig:BSA_sample}) that have no active TTBs, further facilitating structured pruning of their associated weights.
Fig.~\ref{fig:Stratify_vis} further shows the joint effects of stratification and \BSA. 

\begin{figure}[htbp]
    \captionsetup{labelfont={color=black}}
    \centering
    \includegraphics[width=0.99\columnwidth, clip, trim={0.5cm 0cm 0.5cm 0cm}]{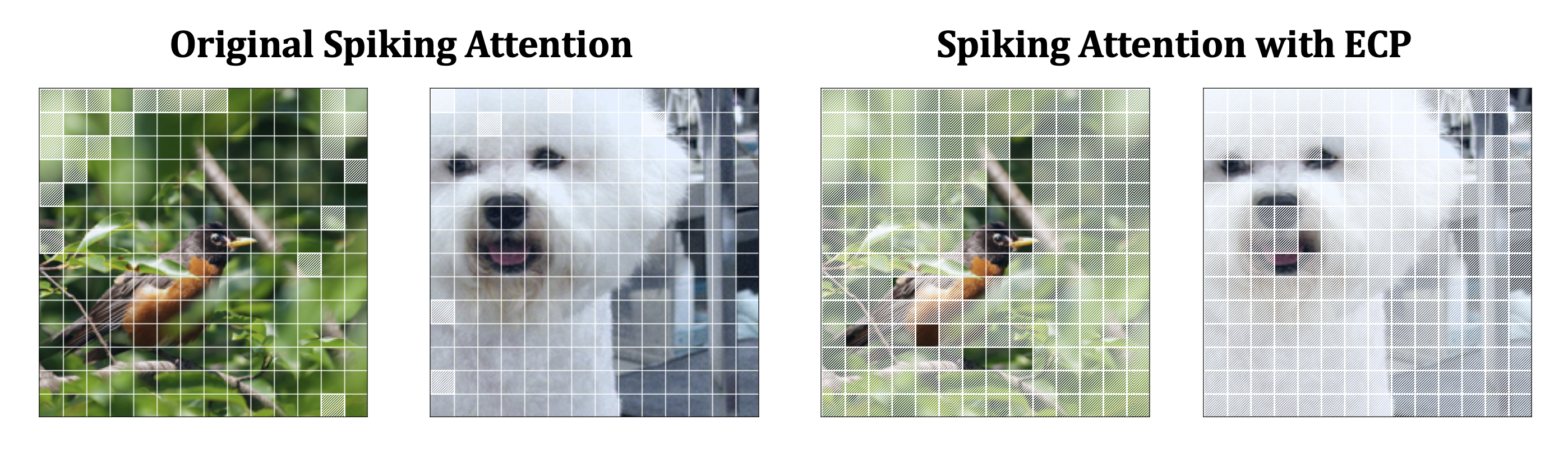}
    \caption{The impact of \ECP\space on the computed attention.}
    \Description{The impact of \ECP\space on the computed attention.}
    \label{fig:ECP_VIS}
\end{figure}

\section{Proposed {\arch} Architecture}
\begin{figure*}[ht]
    \centering
    \includegraphics[width=0.99\textwidth]{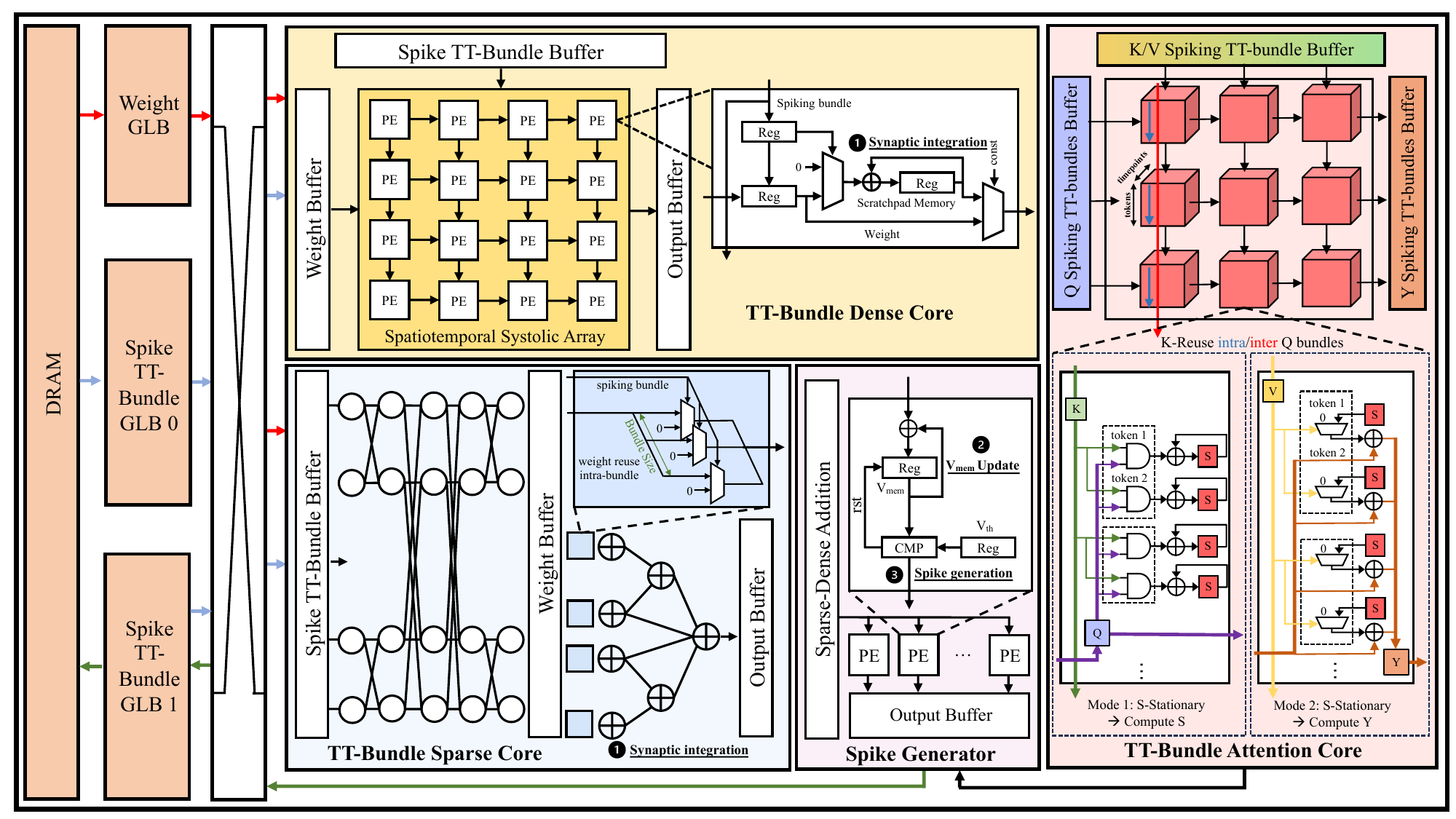}
    \caption{The overall heterogeneous {\arch} Architecture.}
    \Description{}
    \label{fig:Overall_Arch}
\end{figure*}

\subsection{Error-Constrained TT Bundle Pruning (\ECP)}\label{subsection:ecp_algo}

The performance of spiking transformers critically relies on the computation of self-attention maps, whose complexity is quadratic in the number tokens $N$. As $N$ increases,  the computational overhead of  using spiking queries, keys and values to compute increasingly larger spiking attention maps and subsequent computation based on (~\ref{eq:attention}) becomes a dominant bottleneck. 
Models trained over ImageNet, for example, may have $N = 192$ tokens and a fewer number of features, i.e.,  $D = 128$, making the complexity of the spiking self attention (SSA) layers dominate that of the MLP/projection layers. The number of tokens can be even much greater in many long-sequence tasks \cite{Dai2019TransformerXLAL}. We have identified that over 30\% of computational overhead comes from SSA layers in typical spiking transformers. 
The proposed Error-Constrained TT Bundle Pruning (ECP) algorithm prunes away binary activations in the queries (Q) and keys (K), which further triggers a large amount of structured computational reduction in the resulting attention map (S), the values (V), and the outputs (Y) as illustrated in Fig.~\ref{fig:ECP_illustration}.

Notably, \ECP\space explores a key property of the spiking self-attention mechanism that is not present in its ANN counterpart for effective pruning while ensuring a well-controlled error bound. In ANNs, bounding the scores (S) given the queries (Q) and keys (K) is difficult because Q and K are continuous-valued floating point numbers. Differently,  we explore the binary nature of the spiking Q and K for a more straightforward bounding of S without computing them. The total number of active bundles $n_{ab}$ in a particular bundle row across all features of the Q tensor can be easily obtained from the active bundle tags. If $n_{ab}$ is less than a threshold $\theta_{p, Q}$, due to the binary nature of $K$, it is certain that any activation in a bundle that is in the  corresponding row of the scores (S) tensor would be less than $\theta_{p, Q}$ per $S = QK^T$. We prune out this row from the Q tensor entirely while limiting  the error to be no greater than  $\theta_{p, Q}$. In practice, the actual pruning error can be lower than $\theta_{p, Q}$ because the K tensor is sparse and its sparsity pattern does not necessarily coincide with that of Q. Moreover, we adopt the same process to prune the K tensor with a user-specified error bound $\theta_{p, K}$.

As illustrated in Fig.~\ref{fig:ECP_illustration}, pruning Q and K tensors in the above manner has a compounding effect. If 20\% and 10\% of the Q and K bundle rows, respectively, remain after the application of \ECP, 80\% of rows and 90\% of columns in the attention map will be pruned away, reducing the overhead of computing $S$ down to 2\%. This further reduces data access to the $V$ tensor and ultimately decreases the writeback overhead of the $Y$ tensor.
Incorporating \ECP\space into training does not necessarily degrade model accuracy.  
In many cases, \ECP\space can slightly improve model accuracy by acting as a denoising mechanism. Fig.~\ref{fig:ECP_VIS} visualizes the attention maps from the final transformer block of a spiking transformer model trained on ImageNet-100, illustrating how \ECP\space enhances focus on important regions of an input image.

Fig.~\ref{fig:Overall_Arch} shows the proposed heterogeneous {\arch} architecture, which consists of a hierarchical memory system and three main processing cores, targeting the three computational bottlenecks identified in Section~\ref{sec:complexity}. The TT-Bundle Sparse Core and  TT-Bundle Dense Core are dedicated to computing synaptic input integration in the MLP and linear projection layers. The outputs from the two cores are combined by the sparse-dense addition module within the spike generator to compute the final membrane potential $V_{mem}$ for each spiking neuron, which is then used to conditionally generate output spikes.
The TT-Bundle Attention Core is an efficient engine for accelerating spiking self-attention layers. The output from the TT-Bundle Attention Core is fed into the Spike Generator to produce spike-form attention outputs per (\ref{eq:lifattout}).

\subsection{Motivation of the Proposed Architecture}
{\arch} is designed to address the following key issues. 

\textbf{Stratified  workloads for heterogeneous processing.}
\arch\space stratifies sparse and dense workloads for processing on the heterogeneous cores. Fig.~\ref{fig:Stratify}(a) shows that typical spiking spatiotemporal workload is a mix of dense and sparse spiking activities with different sparsity levels on each feature dimension, prohibiting efficient processing. Thanks to our bundle sparsification, we can stratify the workload into dense and sparse parts, as visualized in Fig.~\ref{fig:Stratify_vis}, to be dispatched to the dense and sparse core, respectively, as in Fig.~\ref{fig:Stratify}(b). This significantly improves hardware utilization and avoids scheduling difficulties and inefficiencies that arise from using sparse cores for dense computations, and vice versa.

\textbf{Structured weight reuse based on TTBs.} Packing the workload into TTBs supports multi-bit weight reuse and avoids repeated weight accesses when dealing with irregular spiking patterns on the dense and sparse cores, and allows for key/query data reuse on the spiking attention core. 

\textbf{Tailored core for spiking attention computation.} The multi-bit self-attention computation in ANN-based transformers is expensive and the hardware accelerators bulit on them do not exploit the unique properties of spiking self attention.
To efficiently compute spiking self-attention maps and outputs from binary spike inputs in (\ref{eq:attention}), we propose a reconfigurable systolic array, comprising AND gates, multiplexers, and accumulators as a customized attention core that eliminates the need for costly multi-bit multipliers and significantly reduces area and power overhead.
Furthermore, we employ the feature-first tiling to realize an optimized score-stationary dataflow to reduce data movements of multi-bit attention scores while flowing the binary query/key/value data onto the array. 

\begin{algorithm}[htbp]
\SetAlgoLined
\caption{Stratifying spiking bundles across different input features.}\label{alg:STRA}
\KwIn{Spiking bundled workload: $X \in \mathbb{R}^{B\times D}$, Weight: $W \in \mathbb{R}^{D\times D}$, total number of bundles: $B$, total number of input features: $D$, stratifying column-sparsity threshold: $\theta_{s}$}
\KwOut{Floating low-density spiking bundles $X_{S}$ and the associated weight $W_{S}$, sinking high-density spiking bundles $X_{D}$ and associated weight $W_{D}$}
{\textbf{parallel} }\For{\(i=0\); \(i < D\); \(i\texttt{++}\)}{
    \If{$\sum_{j=0}^{B-1}||S_{j,i}||_{0} > \theta_{s}$}{
        {\textbf{append} $i$ \textbf{to} $R_{D}$; //store dense feature indexes to access permuted weights and spiking bundles}
    }\Else{
        {\textbf{append} $i$ \textbf{to} $R_{S}$; //store sparse feature indexes to access permuted weights and spiking bundles}
    }
}
{$X_{D}$, $W_{D}$ $\leftarrow$ $X_{:, \in R_D}$, $W_{\in R_D, :}$; //routed to dense TTB core}\;
{$X_{S}$, $W_{S}$ $\leftarrow$ $X_{:, \in R_S}$, $W_{\in R_S, :}$; //routed to sparse TTB core}\;
\KwRet {$X_{D}$, $X_{S}$, $W_{D}$ and $W_{S}$}
\end{algorithm}

\subsection{Stratify Sparse and Dense Workload}
\begin{figure}
    \centering
    \includegraphics[width=\columnwidth]{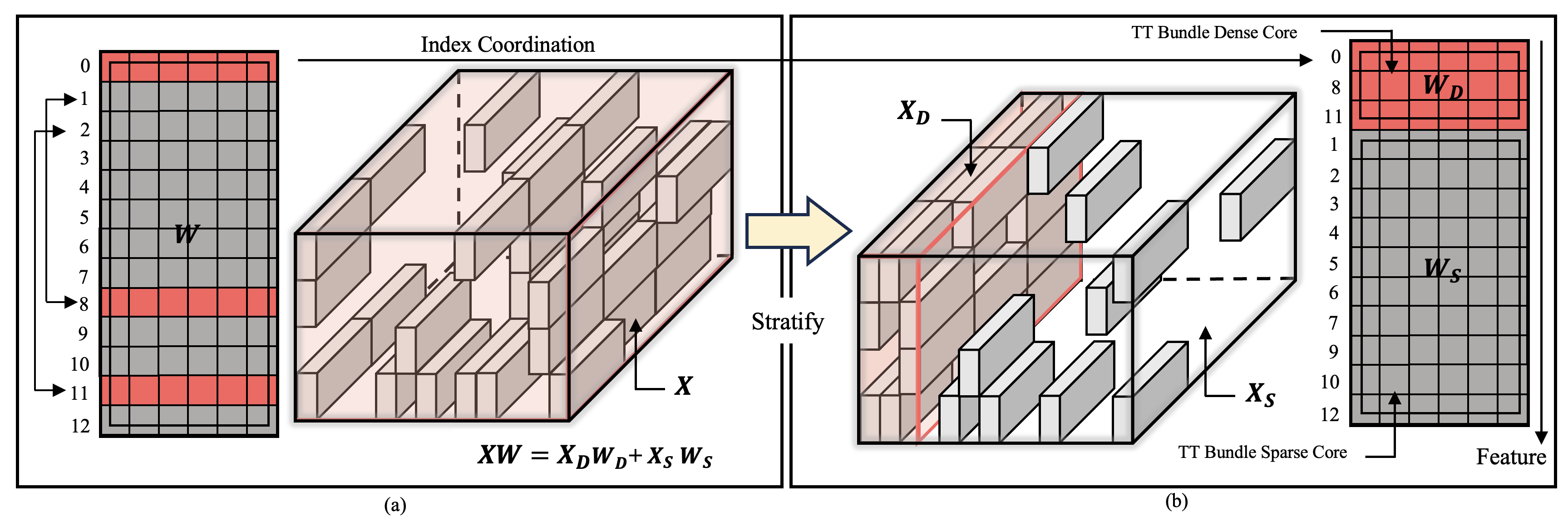}
    \caption{(a) The original spiking  workload $X$. (b) The stratified dense workload $X_{D}$ and sparse workload $X_{S}$; weight indices are coordinated to dispatch $W_{D}$ and $W_{S}$ onto the TT-Bundle dense/TT-Bundle core, respectively.}
    \Description{}
    \label{fig:Stratify}
\end{figure}

Fig.~\ref{fig:Stratify} illustrates how the spiking workload is stratified into the dense and sparse parts, with the feature indices recorded to align rows in the weight matrix so that the Dense TT-Bundle Core processes $X_{D}W_{D}$ and the Sparse TT-Bundle core computes $X_{s}W_{s}$. 
As detailed in Alg.~\ref{alg:STRA}, the stratifier compares the number of active TTBs for each input feature against a stratification threshold to classify the feature to be either ``dense" or ``sparse". A feature index buffer is employed to track the indices of both sparse and dense features. Subsequently, the coordinated weights and input bundles are fed onto the dense/spare core for processing.

\subsection{Dedicated Dense and Sparse Cores}
\textbf{TT-Bundle Dense Core.} 
The stratified dense workload in  the MLP and projection layers is directed to the TT-Bundle Dense Core, which employs an output-stationary architecture, reminiscent of a TPU-like systolic array, as depicted in Fig.~\ref{fig:Overall_Arch}. Each Token-Time bundle is assigned to a distinct processing element (PE). These PEs process and pass spiking bundles in a top-to-bottom sequence. Concurrently, the coordinated weights are passed from left to right across the array, where each PE column  computes a specific output feature. In this process, weight data is repeatedly reused within a bundle in an intra-bundle reuse manner, and reused by all PEs in the same row in an inter-bundle reuse manner. Leveraging the binary nature of spiking inputs, each PE executes ``Select ACcumulation" (SAC) operations to effectively multiply synaptic weights with spiking inputs. A ``SAC'' operation is efficiently implemented by one MUX and one accumulator.  The partial sums are stored in PEs' local registers. Upon completion of the computation assigned to the PE array, the partial sums of synaptic integration are buffered into the output buffer, awaiting merging with the results from the TT-bundle sparse core.

\textbf{TT-Bundle Sparse Core.} 
Working in parallel with the TT-Bundle Dense Core, the TT-Bundle Sparse Core processes the sparse part ($X_{s}W_{s}$) of spiking synaptic integration. We design our TT-bundle sparse core by adopting a SIGMA-like architecture \cite{SIGMA} to efficiently handle irregular sparsity patterns, thanks to its flexible and configurable distribution and reduction network. 
The core utilization is enhanced considerably by the proposed \BSA\space training algorithm presented in Section~\ref{sec:ba_training}, which improves the network's structured bundle-level sparsity. In addition, this core facilitates multi-bit weight data reuse when processing different tokens and time points within a bundle as clustered firing activities may take place in the bundle.   
Finally, we merge and accumulate the partial sums of the synaptic integration streaming out of the sparse and dense cores by performing sparse-dense addition in the Spike Generator. The Spike Generator maps neurons to be processed onto different PEs, where an output spike is conditionally generated if its $V_{mem}$ exceeds the threshold voltage $V_{th}$ at certain time point.

\subsection{Dedicated TT-Bundle Attention Core} \label{subsection:attention_core}
Attention computation in ANN transformers is costly. It involves a sequence of multi-bit multiplications between queries (Q) and keys (K), non-local softmax operations, and multi-bit multiplications of the resulting scores (S) and values (V). 
As described in Section~\ref{sec:spikingattention},  the spiking attention mechanism employs binary tensors Q, K, and V. This, coupled with the elimination of computationally expensive softmax operations, provides a distinct advantage. The proposed attention core efficiently computes spiking attentions by leveraging the binary nature of the Q/K/V data through two-step operations. The first step computes an accumulated attention map by multiplying $Q$ with $K^T$ to produce attention score $S$; the second step is to compute the output $Y$ by multiplying $S$ with $V$. Thanks to our \ECP\space algorithm, there only exist a limited number of $Q/K/V$ bundles that need to be loaded and processed. This selective processing allows the attention array to be dynamically allocated,  solely executing the essential computations.

The attention core utilizes a systolic array while employing optimized dataflows and data-reuse schemes. As shown in Fig.~\ref{fig:Overall_Arch},  each PE has two operation modes. Mode 1 utilizes the flowing $Q$ and $K$ data to compute $S$ with an S-stationary dataflow. Mode 2 utilizes the stationary $S$ and flowing $V$ to compute $Y$. In each PE, the workloads at different time points are mapped onto different groups of logic gates circled by the dotted line.  Multiple groups simultaneously process the same set of tokens for different time points. 

\textbf{Mode 1.} We configure the core into multiple \textbf{A}nd-\textbf{AC}cumulate (AAC) units to efficiently compute the accumulated attention on different tokens at each time point, and store the resulting partial sums in the S registers. \underline{S-stationary dataflow:} Because the attention score $S$ has a higher bit width, ranging from 6 to 10 model-dependent bits, than the binary spikes in Q and K, we adopt an S-stationary dataflow to minimize data movement. This approach involves the directional flow of $Q$ bundles from left to right, coupled with the streaming of $K$ bundles from top to bottom across the array. This setup executes ``AND'' operations between binary queries ($Q$) and keys ($K$), followed by the accumulation of partial sums in the $S$ register within each PE. The process continues until the computation of $S$ is completed for all features. \underline{K-reuse with intra/inter-Q bundles:} K bundles tend to have a higher token sparsity than Q, especially after our error-constrained bundle pruning.
We reuse one set of K-tokens that correspond to the same spatial location across multiple time points to operate with $B_{n}$ tokens from one Q bundle in each PE. Specifically, at a given processing time, each PE contains one Q bundle, and the set of K-tokens is reused for the computation with all tokens in that Q bundle in an intra-Q-bundle manner.  Across various Q bundles mapped onto the PEs in the same column, the K-token set is reused again in an inter-Q-bundle manner.

\textbf{Mode 2.}
The attention core is configured into \textbf{S}elect-\textbf{AC}cumulati
on (SAC) units based on the S-stationary dataflow when computing $Y$. \underline{V-reuse with intra/inter-S bundles:} As shown in Fig.~\ref{fig:Overall_Arch}, we maintain $S$ in the local registers and load $V$ from top to bottom, and flow the computed partial sums $Y$ from left to right. In each ``SAC''operation, the binary $V$ input selects the right $S$ data to be accumulated into the partial sum, which is stored in the $Y$ register. 
In a fashion that is similar to the K-reuse described above, V is reused in both an intra- and inter-S bundle manner. In each PE, $V$ is reused for processing different S tokens for multiple time points to facilitate the intra-S bundle reuse. The same $V$ data is reused again in an inter-S bundle manner across the PEs in the same column.  When $Y$ is read out, it is aggregated into the partial sum maintained within the $Y$ TT-bundle buffers. Upon completion of $Y$'s computation, with $Y$ represented in an integer format, a shifter is employed to adjust the scale of $Y$ per (~\ref{eq:attention}) based on a power-of-two scaling factor $s$. The processed $Y$ is then fed into the spiking generator, and the final binary  attention output is  stored back into the TTB GLBs.

\section{Evaluation}
\subsection{Evaluation Setup}\label{subsection:Eval_setup}
\textbf{Model Training and Datasets.} We develop a  training flow based on Pytorch, integrating the proposed Bundle-Sparsity Aware Training (\BSA) and Error-Constrained TTB Pruning (\ECP) algorithms. Several spiking transformer models are trained on widely adopted image recognition datasets including CIFAR10, CIFAR100, ImageNet-100\cite{deng2009imagenet}, the neuromorphic dynamic vision sensor (DVS) dataset DVS-Gesture-128 \cite{amir2017low}, and speech command recognition dataset Google Speech Command V2 \cite{warden2018speech}.

To evaluate the scalability of our proposed {\arch} architecture, we train multiple spiking transformers with different model architectures, following the settings in Tab.~\ref{table:trans_arch}. In the case of CIFAR10 and CIFAR100, each image is sized at $32 \times 32$, which is segmented into 64 ($N$) tokens, with each token representing a $4\times 4$ pixel area. The feature size ($D$) is set to 384. The fact that $D >> N$  renders the MLPs and linear projection layers as the dominant contributors to computational complexity. An ImageNet-100 image has a resolution of $224\times 224$ pixels per channel. We split each image into 196 tokens, each representing a $16\times 16$ pixel area. The feature size is set to 128, making the attention layers the most dominant source of computational complexity because $N > D$. For the neuromorphic dataset DVS-Gesture-128, the visual input at a time step comprises $128\times 128$ pixels, which is divided into 64 tokens, each sized at $16\times 16$. We use a batch size of 256 for 300 training epochs on the CIFAR and ImageNet-100 datasets, and a batch size of 64 for 100 epochs on the DVS-Gesture-128 dataset. Advances in training algorithms have resulted in high-accuracy SNNs operating on a reduced number of time steps. Different from the prior work \cite{PTB, Spinalflow, skipper} that employ 100 to 300 time steps, we limit the number of time steps to be from $4$ to $20$. This reduction  contributes to a lower training cost and aligns with the trend in developing high-performance spiking neural models \cite{TSSLBP, spikformer} for low-latency processing.

In implementing \BSA, we configure the parameter $\lambda$ differently for various datasets: $1$ for CIFAR10, $0.5$ for CIFAR100, $0.3$ for ImageNet, and $1.0$ for DVS-Gesture-128. This choice allows us to strike a balance between bundle sparsity and accuracy tailored for each dataset. For error-constrained pruning of queries and keys in attention layers, we set the bundle pruning thresholds to 10 for models trained on DVS-Gesture-128 and 6 for other models without compromising accuracy.

\begin{table}[htbp]
\setlength{\tabcolsep}{2.5pt}
    \centering
    \small
    \caption{Spiking transformer architectures for three static datasets and one dynamic dataset}
    \begin{tabular}{c c c c c c c}
    \toprule
        Type & Model & Dataset  & \makecell[c]{Blocks\\($B$)} & \makecell[c]{Timesteps\\($T$)} &  \makecell[c]{Tokens\\($N$)} &  \makecell[c]{Features\\($D$)} \\
    
    \midrule
    
        \multirow{3}{*}{Static} & Model 1 & CIFAR10      & 4  & 10   & 64    & 384\\
                                & Model 2 & CIFAR100     & 4  & 8    & 64    & 384\\
                                & Model 3 & ImageNet100  & 8  & 4    & 196   & 128\\
    \midrule
    
        \multirow{1}{*}{Dynamic}& Model 4 & DVS-Gesture  & 2  & 20   & 64    & 128\\
    \midrule
        \multirow{1}{*}{\textcolor{black}{Language}}& \textcolor{black}{Model 5} & \textcolor{black}{Google SC}  & \textcolor{black}{4}  & \textcolor{black}{8}   & \textcolor{black}{256}    & \textcolor{black}{384}\\
    \bottomrule
    \end{tabular}
    \label{table:trans_arch}
\end{table}

\textbf{Baselines and Evaluation Metrics.} \underline{Baselines}: To benchmark the proposed {\arch} architecture, we compare it with an edge GPU (NVIDIA Jetson Nano) and Parallel Time Batching (PTB), a recent competitive SNN accelerator architecture\cite{PTB}. For a fair comparison, PTB and the proposed {\arch} are configured to have the same number of PEs with each PE possessing the same amount of register and compute resources, resulting in  nearly identical area and operational power when synthesized using a commercial 28nm process design kit (PDK). 
\underline{Evaluation Metrics}: We consider chip area, power, latency, and energy dissipation. 

\textbf{Modeling of Architecture, Latency, Energy Dissipation, and Synthesis.} We build an analytic cycle-accurate heterogeneous-core architecture simulation environment to support all unique features in spiking transformers. It traces data movement between the heterogeneous cores and hierarchical memories for assessing latency and energy dissipation. We follow the standard practice to employ a three-level memory hierarchy for memory-intensive transformer computations \cite{ViTCoD, Taskfusion}. Similar to many other analytic models, each level of memory is double-buffered to hide latency and equally partitioned to separately store different types of data. 

We use CACTI 7.0 \cite{CACTI7} to estimate energy dissipation of the weight global buffer (GLB) and spiking TT-bundle GLBs. The weight GLB is a 144KB SRAM with 512-bit read/write ports.  The ping-pong spiking TT-bundle GLB0/GLB1 with each being 12KB can be equally partitioned for storage of binary spiking Q/K/V/Y. The employed DRAM has a DDR4-2400 memory bandwidth of 76.8GB/s and power consumption of 323.9mW at a core frequency of 500MHz.  In \arch, the TT-bundle sparse core consists of up to 128 parallel TT-bundle processing units. Both the TTB dense core and  TTB attention core consist of 512 PE units, processing up to 32 output features and 16 TT-bundles in parallel. Each TTB unit can reconfigurablely process up to 10 spikes in one cycle. The spike generator  may  process up to 512 neurons in parallel.
We use the open-source SIGMA simulator STONNE \cite{STONNE21} for cycle-accurate simulation of the TTB sparse core. 

The RTL implementation of the  {\arch} accelerator is synthesized using a commercial 28nm CMOS technology, resulting to  a die area of
$2.96 mm^2$, a peak power dissipation of $627 mW$, and a clock rate of 500 MHz. In comparison,  the synthesized baseline PTB accelerator has a chip area of $2.80 mm^2$ and a peak power dissipation of $606.9 mW$. 

\subsection{Overall Latency and Energy Evaluation}\label{subsection:overall_eval}
\begin{figure}[ht]
    \centering
    \includegraphics[width=\columnwidth, clip, trim={0.1cm 1.2cm 7cm 0.1cm}]{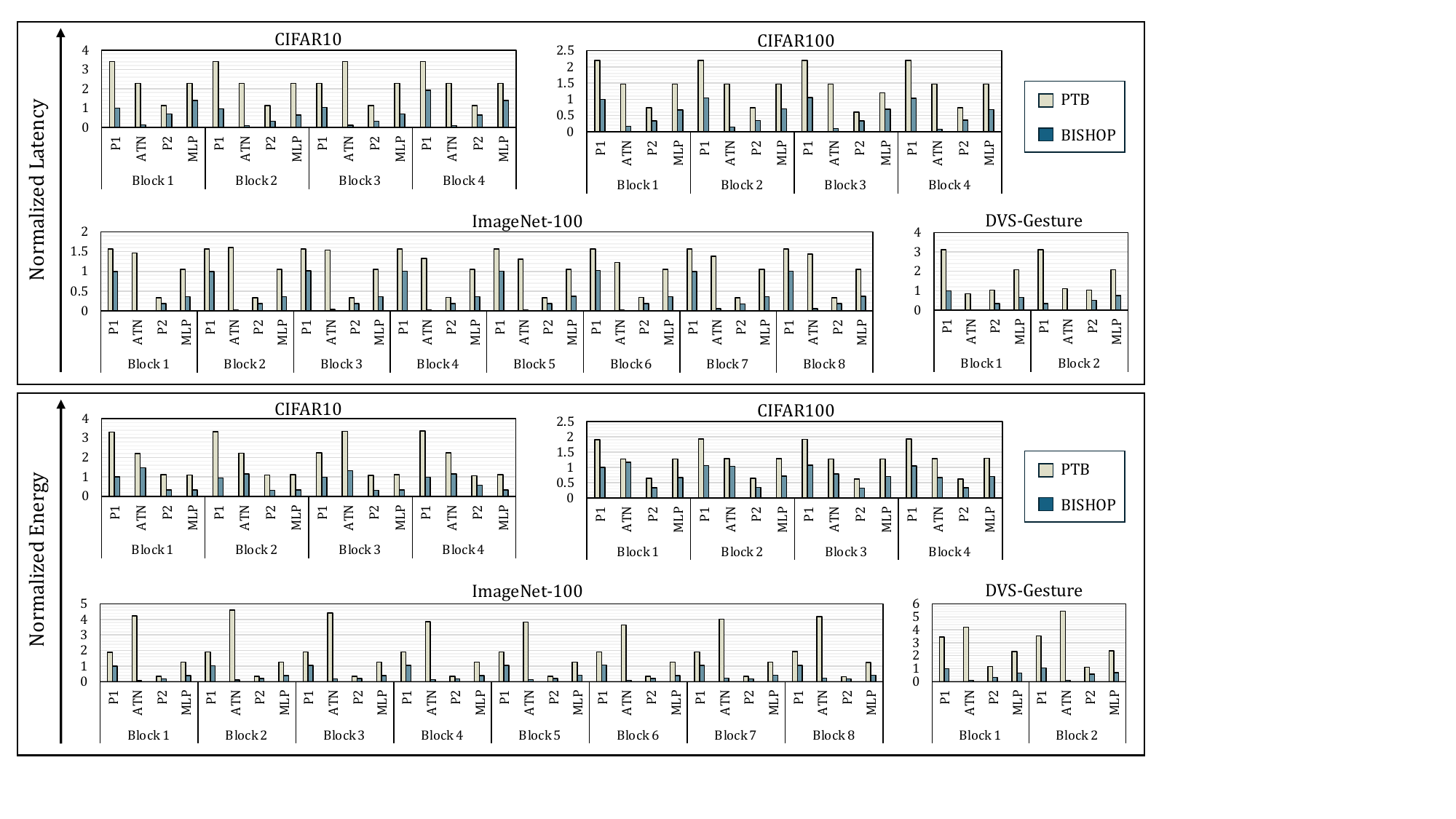}
    \caption{Normalized latency and energy comparison of {\arch} with PTB \cite{PTB} when accelerating the same spiking transformers trained on CIFAR10, CIFAR100, ImageNet-100 and DVS-Gesture-128. The latency and energy consumption are normalized by those of the first projection layer of the first block in {\arch}, respectively. P1, ATN, and P2 indicate the $Q/K/V$ linear projection layer, spiking self-attention layer, and $O$ linear projection layer within the spiking self-attention block, respectively; MLP indicates the spiking MLP layer.}
    \Description{}
    \label{fig:layer_wise_all}
\end{figure}

\begin{figure}[htbp]
    \captionsetup{labelfont={color=black}}
    \centering
    \includegraphics[width=\columnwidth, clip, trim={0cm 0cm 0.8cm 0cm}]{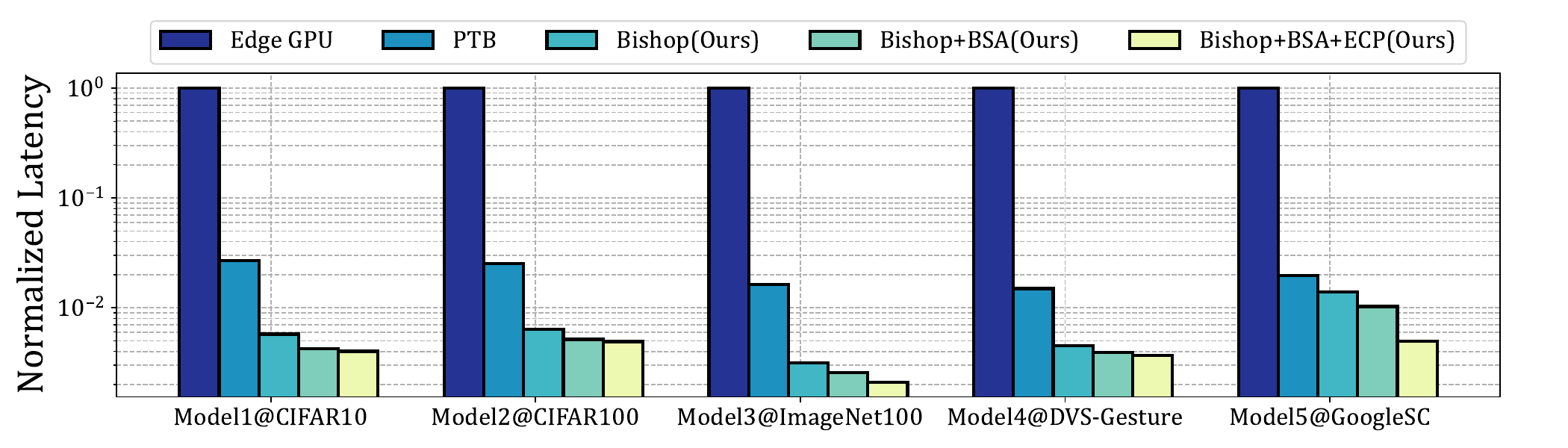}
    \caption{Evaluation on end-to-end normalized latency reduction.}
    \Description{}
    \label{fig:E2E_latency}
\end{figure}

\begin{figure}[htbp]
    \captionsetup{labelfont={color=black}}
    \centering
    \includegraphics[width=\columnwidth, clip, trim={0cm 0cm 0cm 0cm}]{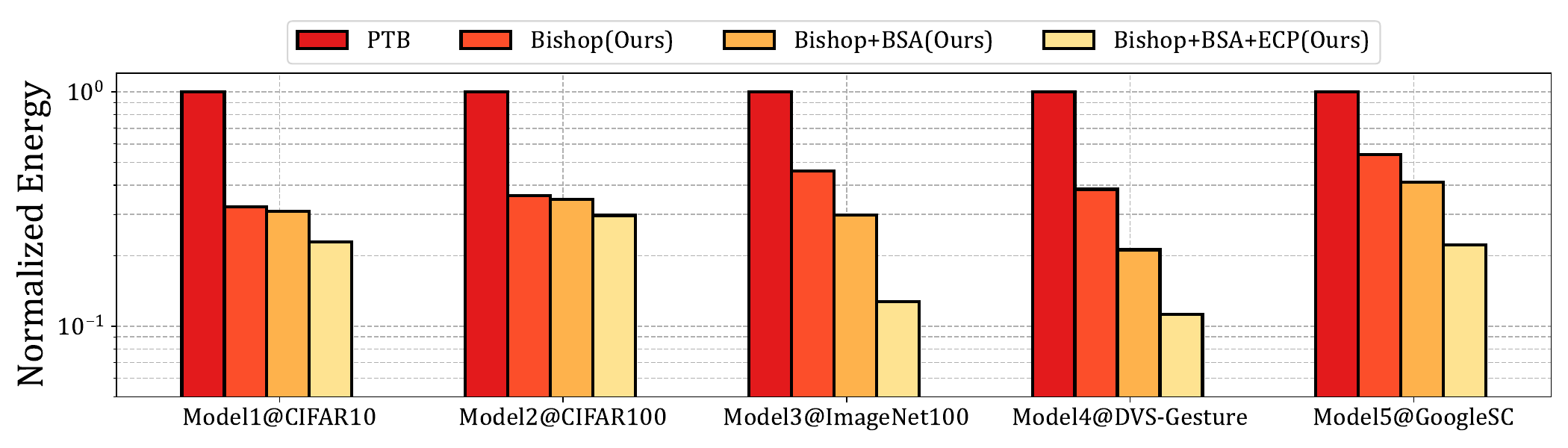}
    \caption{Evaluation on end-to-end normalized energy consumption reduction.}
    \Description{}
    \label{fig:E2E_energy}
\end{figure}

Fig.~\ref{fig:E2E_latency} and Fig.~\ref{fig:E2E_energy} show the end-to-end performance of our \arch\space accelerator and the two baselines. Fig.~\ref{fig:layer_wise_all} shows more detailed layer-wise speedup and energy saving of  {\arch} in comparison with PTB \cite{PTB}.
Across four pre-trained models, \arch \space on average achieves $299\times$ and $5.91\times$ speedups compared with the edge GPU and PTB~\cite{PTB}, respectively, while reducing energy by $6.11\times$ over PTB. Notably, these significant latency and energy reductions are achieved without comprising model accuracy.

On CIFAR10, \arch\space speeds up the edge GPU and PTB by $173.9\times$ and $4.68 \times$; \arch+\BSA\space has a $238.6\times$ and $6.37\times$ speedup over the edge GPU and PTB; \arch+\BSA+\ECP\space has a speedup of $249.4\times$ and $6.71\times$ over the edge GPU and PTB, respectively. 
On CIFAR100, \arch\space offers a $156\times$ and $3.95 \times$ speed up over the edge GPU and PTB; \arch+\BSA\space provides a $193.9\times$ and $4.90\times$ speedup over the edge GPU and PTB; The speedups are $203.3\times$ and $5.14\times$, respectively, over the edge GPU and PTB by using \arch+\BSA+\ECP.
On ImageNet-100, \arch\space has a $317.6\times$ and $5.17 \times$ speed up over the edge GPU and PTB; \arch+\BSA\space has a $389\times$ and $6.34\times$ speedup over the edge GPU and PTB; \arch+\BSA+\ECP\space speeds up the edge GPU and PTB by $474.8\times$ and $7.73\times$, respectively.
On DVS-Gesture-128, \arch\space has a $221\times$ and $3.30 \times$ speed up over the edge GPU and PTB; \arch+\BSA\space increases the speedups to $254.9\times$ and $3.81\times$; \arch+\BSA+\ECP\space has $271.76\times$ and $4.06\times$ speedup over the edge GPU and PTB, respectively.

For model 5 evaluated on the language task Google Speech Command \cite{warden2018speech}, 
\arch\space achieves a speedup of $72.2\times$ over the edge GPU and $1.43\times$ over PTB. With \BSA, the speedups increase to $97.33\times$ and $1.92\times$, respectively. Further incorporating \ECP\space boosts the speedups to $202.42\times$ over the edge GPU and $4.0\times$ over PTB.

\subsection{Evaluation of \arch\space Algorithms}\label{subsection:algo_eval}

\begin{figure}
    \captionsetup{labelfont={color=black}}
    \centering
    \includegraphics[width=\columnwidth, clip, trim={2cm 3cm 11cm 0cm}]{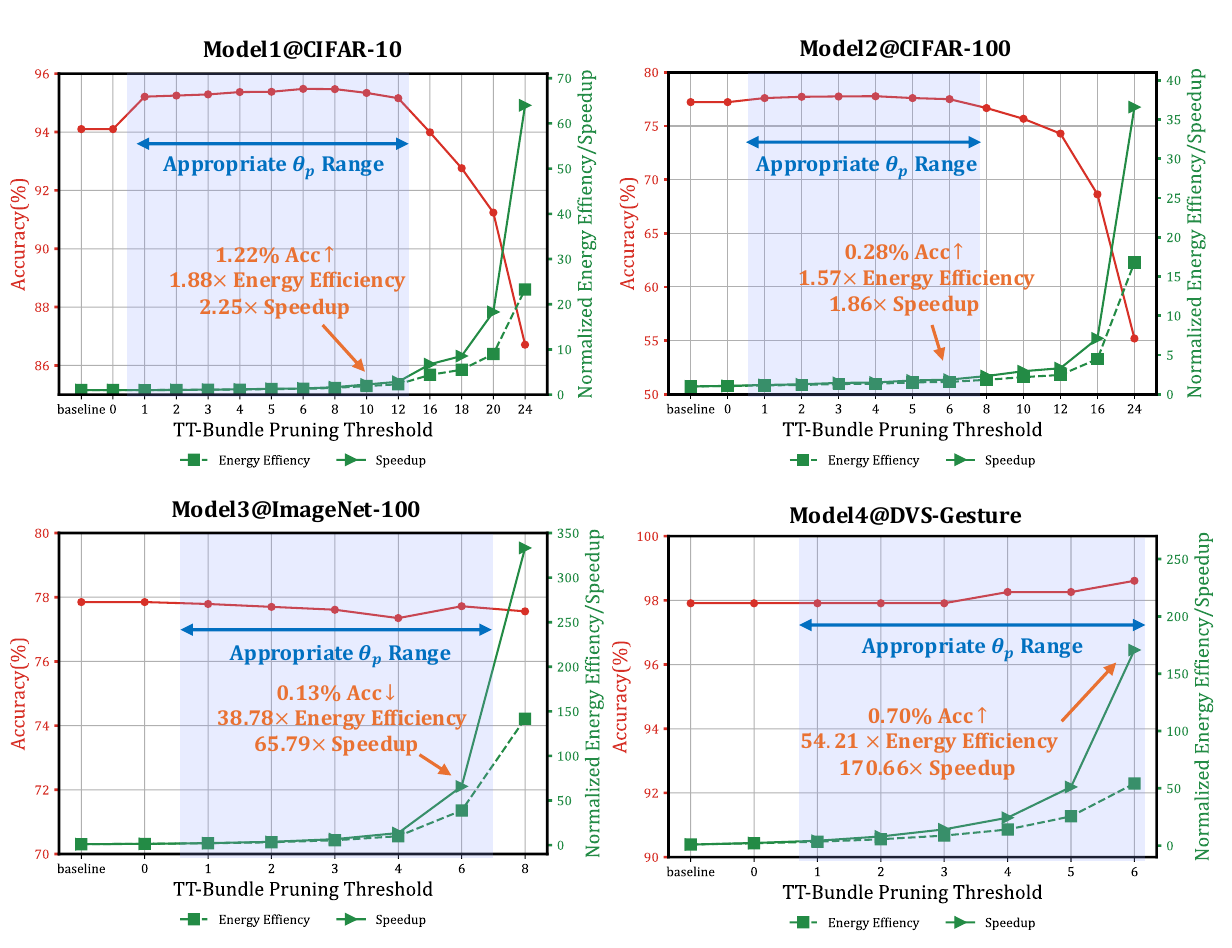}
    \caption{Accuracy v.s. normalized energy efficiency and speedup of the spiking self-attention layers of four spiking transformers under different \ECP\space pruning thresholds.}
    \Description{}
    \label{fig:pruning_acc}
\end{figure}

Fig.~\ref{fig:pruning_acc} quantifies the impact of the pruning threshold of \ECP\space on the energy efficiency and speedup of spiking self-attention layers. Across a range of pruning thresholds, a significant proportion of bundles are removed from the Q/K/V tensors without causing any substantial drop in model accuracy. In some cases, pruning even improves accuracy by introducing a denoising effect.

\begin{figure}[htbp]
    \captionsetup{labelfont={color=black}}
    \centering
    \includegraphics[width=\columnwidth, clip, trim={0cm 0cm 0cm 0cm}]{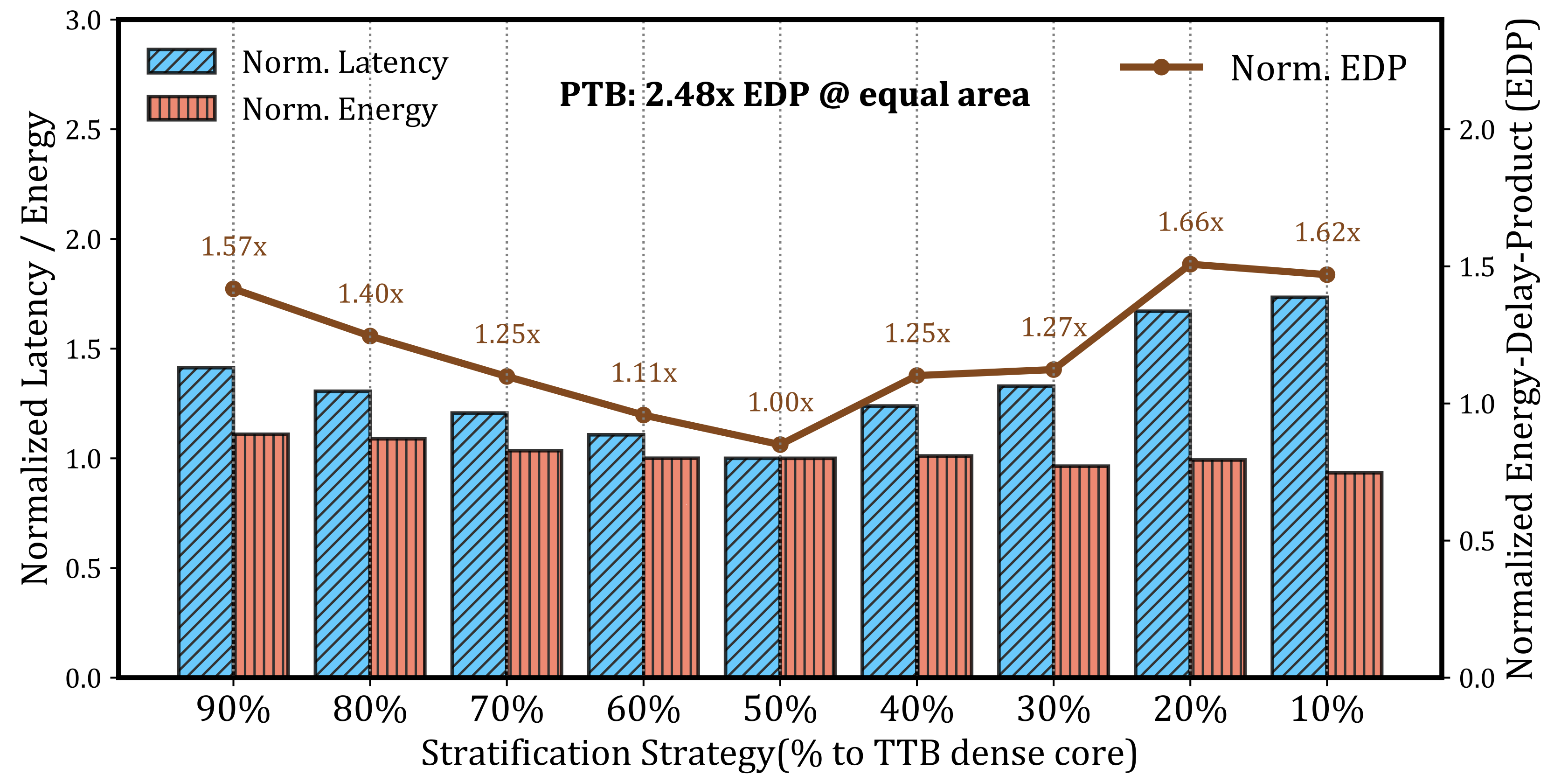}
    \caption{Impact of stratifacation strategies on the energy, latency and EDP of Model 3 trained on ImageNet-100. Different stratification strategies yield varying stratification thresholds ($\theta_{s}$) to achieve a targeted dense-to-sparse core token split ratio.}
    \Description{}
    \label{fig:sweep_theta_s}
\end{figure}


\begin{figure*}[htbp]
    \captionsetup{labelfont={color=black}}
    \centering
    \includegraphics[width=0.95\textwidth, clip, trim={0cm 0cm 0cm 0cm}]{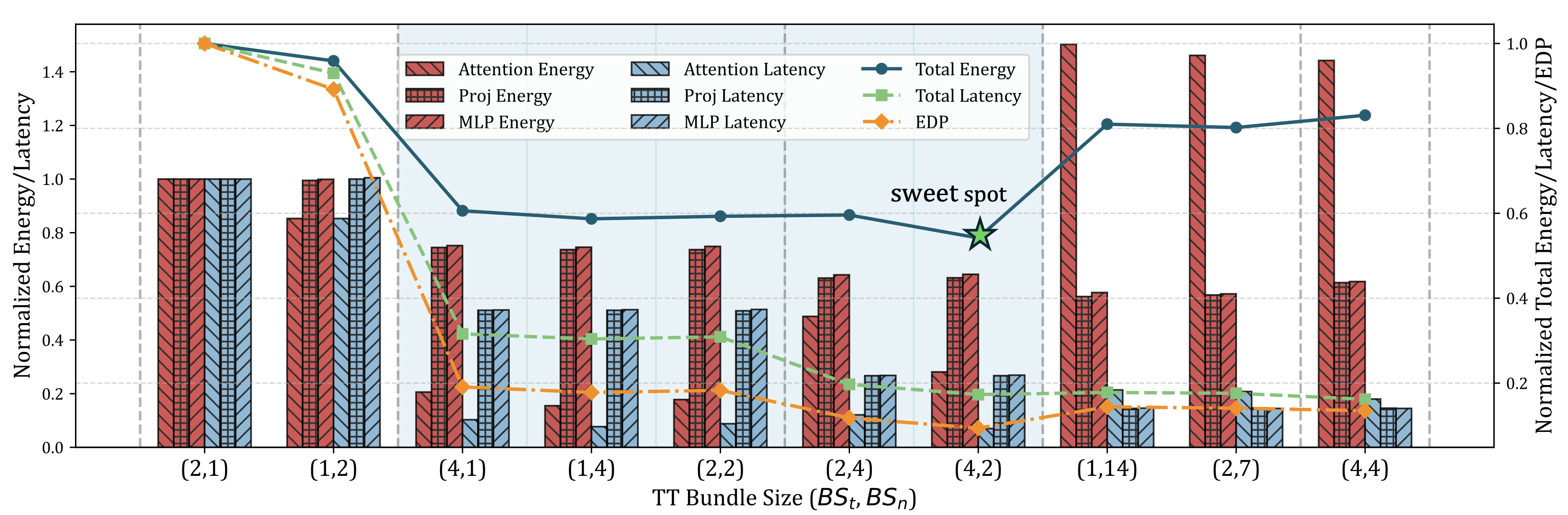}
    \caption{Sensitivity analysis of TTB bundle volume $(BS_t, BS_n)$  for Model 3 trained on ImageNet-100.}
    \Description{}
    \label{fig:sweep_bundle_size}
\end{figure*}

Nevertheless, using a large pruning threshold could degrade model performance, hence it should be properly set within an appropriate range to trade off between hardware performance and accuracy. 
For example, with the TTB pruning threshold $\theta_{p}$ set to 8 on CIFAR10, the model accuracy is improved by $1.24\%$ while $28.2\%$ of the Q tokens and $48.0\%$ of the K tokens are pruned away, reducing the attention map computation by $62.7\%$ and memory access by $38.1\%$. The total energy and latency are reduced to $0.52\times$ and $0.44\times$, respectively.
For CIFAR100, the model's accuracy is improved by $0.53\%$ while keeping $93.2\%$ and $55.1\%$ of the tokens in the queries and keys, reducing energy dissipation to $0.65\times$ and latency to $0.42\times$, respectively. 
On ImageNet-100, \ECP\space leads to a slight accuracy drop of 0.13\% while retaining only $10.7\%$ of the query tokens and $9.7\%$ of the key tokens. This drastic pruning reduces attention map computation overhead down to $1.04\%$, reducing energy to $0.03\times$ and latency to $0.02\times$ for the self-attention layers. On DVS-Gesture-128, the model's performance is enhanced by $0.7\%$ while pruning the vast majority of tokens. Only $8.0\%$ of the Q tokens and $5.49\%$ of the K tokens remain. On average, the spiking Q tokens can be pruned away by $51.71\%$, reaching up to $92\%$ in some cases.  Similarly, the average pruning of the spiking K tokens can reach $67.77\%$ with a peak reduction of $97.51\%$. Consequently, on average, only $15.5\%$ of the computation is performed, leading to an average $83.76\%$ decrease in energy consumption and $43.92\%$ decrease in latency for the computations of the self-attention layers.

\subsection{Evaluation of {\arch} Hardware Accelerator}\label{subsection:hardware_eval}
We compare {\arch} and PTB \cite{PTB} purely from a hardware architectural perspective by removing the proposed  \BSA\space and \ECP\space algorithms and using the ImageNet-100 dataset. 

We first demonstrate the effect of the heterogeneity of {\arch}. The spiking transformer trained on the ImageNet-100 dataset has an average of 20\% sparsity level across all layers.  The stratifier in {\arch} directs 50\% of the workload to the TT-bundle dense core, and dispatches the remaining sparser workload to the TT-bundle sparse core. 
For the inference of a single image, on average the dense core takes 1.16ms and consumes 0.29mJ of energy. The sparse core takes 0.53ms while burning 0.038mJ of energy.  In contrast,  if all workload is processed by the dense core as in the case of PTB, the latency and energy increases to 1.83ms and 0.45mJ, respectively. Thus, the heterogeneity of {\arch} offers a 1.39$\times$ speed-up and a 1.57$\times$ energy saving. 

As for spiking attention computation,  the dedicated attention core of  {\arch} reduces latency by 10.7-23.3$\times$ while achieving 1.39-1.96$\times$ energy saving over PTB.

\subsection{Design Space Explorations}\label{sec:DSE}
There are two important architectural hyperparameters, namely, the stratification threshold and TTB bundle volume, which have a large impact on \arch's performance.

\subsubsection{Impact of Stratification Threshold}\label{sec:s_threshold}
Choosing a higher value for the stratification threshold $\theta_s$ shifts more workload from the TTB dense core to the TTB sparse core.
Nevertheless, a large imbalance in workload between the two cores can degrade \arch's performance. 
Using the dense core to process an excessive amount of workloads, which can constitute a large portion of the sparse workloads that cannot be efficiently processed by the dense core, may reduce the utilization of the sparse core and degrade the overall latency of \arch. Conversely, overloading the sparse core reduces its efficiency in processing assigned dense workloads, lowers the utilization of the dense core, and degrades the overall latency. 
In contrast, the stratification threshold $\theta_s$ has a relatively minor effect on the energy dissipation of \arch, since data movement continues to dominate overall energy costs. Nonetheless, shifting more workload to the dense core can result in a slight increase in energy consumption.
As a result, the Energy-Delay Product (EDP) initially decreases and then increases as $\theta_{s}$ increases.

Fig.~~\ref{fig:sweep_theta_s} details the impact of different workload distribution strategies on energy, latency, and EDP. In practice, a near-optimal EDP can be achieved when the stratification threshold $\theta_s$  of each layer is chosen to  approximately balance the workload between the two cores. This results in an EDP improvement of $2.49\times$ compared to the PTB architecture, when realized with an equal chip area in the 28nm technology. In contrast, imbalance between the two cores can degrade the \arch's EDP by up to $1.65\times$. 

\begin{figure}[htbp]
    \centering
    \includegraphics[width=0.99\columnwidth, clip, trim={2cm 4cm 6cm 4cm}]{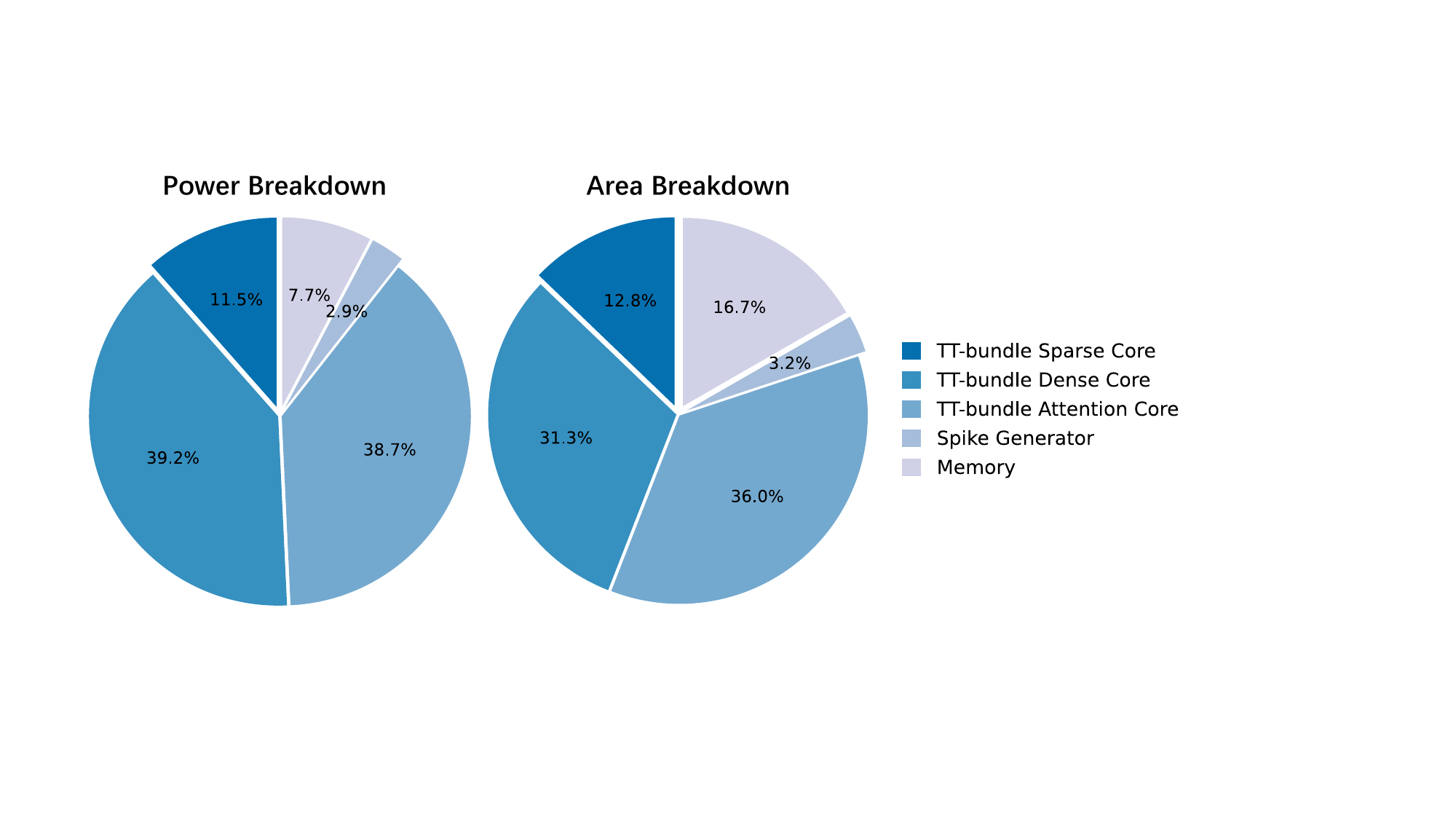}
    \caption{Power/area breakdown of the proposed \arch\space accelerator.}
    \Description{}
    \label{fig:breakdown}
\end{figure}

\subsubsection{Impact of TTB Bundle Volume}
As shown in Fig.~\ref{fig:Bundle_concept}, the TT bundle volume specifies the number of spatiotemporal tokens packed in a bundle and is defined by $BS_t\times BS_n$, where $BS_t$ and $BS_n$ are the temporal and spatial bundle sizes, respectively.
As shown in Fig.~\ref{fig:MultiHead}, the computation workload can be divided into self-attention layers, projection layers and MLP layers. 
Fig.~\ref{fig:sweep_bundle_size} evaluates the impact of the TTB volume on the \arch\space accelerator's energy and latency.

For the attention layers, increasing the bundle volume reduces the effectiveness of the proposed ECP pruning, leading to more bundles with only a small number of active spikes. However, these bundles still need to be transferred from the DRAM to the GLBs and eventually to the attention cores, which can degrade both energy efficiency and latency. On the other hand, when the bundle volume $(BS_t, BS_n)$ is very small, such as $(1,2)$ or $(2,1)$, the intra-bundle and inter-bundle key/value data reuse drops, leading to increased data movement and higher latency.

For the projection and MLP layers, 
initially increasing the bundle volume boosts multi-bit weight data reuse. However, when the bundle volume goes beyond a certain value, such as 14 in Fig.~\ref{fig:sweep_bundle_size}, many idle tokens get bundled into a TTB.
This leads to increased memory traffic and unnecessary processing of idle tokens, resulting in higher energy dissipation, as spiking activation memory access becomes more dominant than multi-bit weight data access. 
For example, when the bundle volume increases from (2,4) to (4,14), spiking activation memory access rises from 13\% to 21.4\% while weight memory access drops from 36.9\% to 16.9\%.
In practice, selecting a bundle volume between 4 and 8, as indicated by the light blue region in Fig.~\ref{fig:sweep_bundle_size},  achieves near-optimal total energy and latency.

\subsection{Area and Power breakdown}\label{subsection: area}
Fig.~\ref{fig:breakdown} shows the area and energy breakdown of the \arch\space accelerator synthesized using a commercial 28nm technology.  Nearly 90\% of the total power and 80\% of the chip area  are consumed by the three major cores. The TTB sparse core, TTB dense core,  and TTB attention core consume 72.2$mW$(11.5\%), 246.1$mW$(39.2\%) and 242.51$mW$(38.7\%), and occupy 0.38$mm^{2}$(12.8\%), 0.92$mm^{2}(31.3\%)$, 1.06$mm^{2}(36.0\%)$ chip area, respectively.
In contrast, the spiking generator array consumes 0.09$mm^{2}(3.2\%)$ area and $18.1mW(2.9\%)$, and the GLBs consume $0.495mm^{2}(16.7\%)$ area and $48.3mW(7.7\%)$, respectively.  The total die area and peak power of the \arch\space accelerator are  $2.96mm^2$ and $627mW$, respectively, similar to  the PTB accelerator ($2.80mm^{2}$, $606.9mW$).

\section{Related Work}\label{sec:related_work}
Among the prior SNN accelerators discussed in Section~\ref{sec:existing_snn}, PTB~\cite{PTB} batches processing of multiple time steps for a given neuron within a time window, allowing multi-bit weight sharing within the  window. 
Several other works have also explored temporally parallel processing of spiking CNN workloads.
While not designed specifically for spiking transformers, 
LoAS~\cite{yin2024loas} explores both the input activation and weight data sparsity for further efficiency improvements, particularly for SNNs with a small number of time steps. Stellar~\cite{mao2024stellar}  accelerates SNNs based on few-spike (FS) neurons instead of LIF neurons.  Importantly, all these works focus on spiking CNNs with homogeneous accelerator architectures while \arch\space
introduces agile spatiotemporal processing and a heterogeneous architecture tailored for complex spiking transformer workloads.  

Like \arch, sparsity has been a key focus in prior work. \cite{ActivityRegularization} proposes a training method to enhance spike-level sparsity, whereas \arch\space maximizes structural TT-bundle-level sparsity, which is more critical for hardware acceleration. While sparse attention mechanisms have been explored in ANN-based transformer accelerators~\cite{lu2021sanger, wang2021spatten}, these methods rely on post-attention computation pruning unlike the proposed Error-Constrained TTB Pruning (ECP). In addition, they do not target binary spiking attention mechanisms.

\section{Conclusions}
We present \arch, the first dedicated hardware accelerator architecture and HW/SW co-design framework for spiking transformers.
{\arch} operates on spiking time-token bundles (TTBs) to minimize weight data access and explore structured bundle-level sparsity.
\arch\space utilizes a bundle sparsity-aware training pipeline to improve structured TTB sparsity, and error-constrained pruning to aggressively trim spiking queries and keys, thereby significantly reducing the overhead of computing large spiking attentions. \arch\space incorporates  a dedicated TT-bundle dense/sparse core,  a dense/sparse workload stratifier, and a dedicated spiking attention core to reduce data movement and boost acceleration efficiency. Extensive experimental studies have demonstrated significant advantages of \arch\space over the prior SNN accelerators. 

\section*{ACKNOWLEDGMENTS}
This material is based upon work supported by the National Science Foundation under Grants No. 1948201 and No. 2310170. 


\bibliographystyle{IEEEtranS}
\bibliography{refs}


\end{document}